%% file: main.tex
\documentclass{article}

\usepackage[preprint]{neurips_2026}

\usepackage[T1]{fontenc}
\usepackage{hyperref}
\usepackage{url}
\usepackage{microtype}
\usepackage{xcolor}
\usepackage{graphicx}
\usepackage{booktabs}
\usepackage{algorithm}
\usepackage{algpseudocode}
\usepackage{amsmath}
\usepackage{amssymb}
\usepackage{amsfonts}
\usepackage{mathtools}
\usepackage{amsthm}
\usepackage{enumitem}
\usepackage[para,online,flushleft]{threeparttable}
\usepackage[capitalize,noabbrev]{cleveref}
\usepackage[printonlyused]{acronym}
\makeatletter
\renewcommand*{\AC@verridelabel}[1]{%
  \@bsphack
  \label{#1}%
  \AC@overriddenmessage rs{#1}%
  \@esphack
}
\makeatother
\setlist{label=\textbullet}
\theoremstyle{plain}
\newtheorem{theorem}{Theorem}[section]
\newtheorem{proposition}[theorem]{Proposition}

\theoremstyle{definition}

\theoremstyle{remark}

\definecolor{myred}{HTML}{C84B31}
\definecolor{myblue}{HTML}{2C5F9E}
\definecolor{mygray}{HTML}{5A5A5A}
\definecolor{mygreen}{HTML}{2F6F4F}
\definecolor{mygold}{HTML}{B07A00}
\definecolor{myteal}{HTML}{2F7A72}

\acrodef{rl}[RL]{Reinforcement Learning}
\acrodef{mdp}[MDP]{Markov Decision Process}
\acrodef{bc}[BC]{behavior cloning}
\acrodef{drol}[DROL]{Dynamic Routing for Offline Reinforcement Learning}
\acrodef{fql}[FQL]{Flow Q-Learning}
\acrodef{imle}[IMLE]{Implicit Maximum Likelihood Estimation}
\acrodef{d4rl}[D4RL]{Datasets for Deep Data-Driven Reinforcement Learning}

\title{Preserve Support, Not Correspondence: Dynamic Routing for Offline Reinforcement Learning}

\author{%
  Zhancun Mu$^{1}$ \quad Guangyu Zhao$^{1}$ \quad Yiwu Zhong$^{1}$ \quad Chi Zhang$^{1,\dagger}$\\
  $^{1}$School of Intelligence Science and Technology, Peking University\\
  \texttt{\{muzhancun,zhaogy24\}@stu.pku.edu.cn}\\
  \texttt{yiwu-zhong@outlook.com \quad chizhang.cz@pku.edu.cn}\\
  \texttt{$^\dagger$ Correspondence to: chizhang.cz@pku.edu.cn}
}

\begin{document}

\maketitle

\begin{abstract}
\input{section/abstract}
\end{abstract}

\input{section/introduction}
\input{section/preliminaries}
\input{section/method}
\input{section/theory}
\input{section/experiments}
\input{section/conclusion}
\input{section/references}
\clearpage
\appendix
\onecolumn

\input{section/appendix_implementation}
\input{section/related_work}
\input{section/appendix_proofs}
\input{section/appendix_results}

\end{document}

%% file: section/abstract.tex
One-step offline RL actors are attractive because they avoid backpropagating through long iterative samplers and keep inference cheap, but they still have to improve under a critic without drifting away from actions that the dataset can support.
In recent one-step extraction pipelines, a strong iterative teacher provides one target action for each latent draw, and the same student output is asked to do both jobs: move toward higher Q and stay near that paired endpoint.
If those two directions disagree, the loss resolves them as a compromise on that same sample, even when a nearby better action remains locally supported by the data.
We propose DROL, a latent-conditioned one-step actor trained with top-1 dynamic routing.
For each state, the actor samples $K$ candidate actions from a bounded latent prior, assigns each dataset action to its nearest candidate, and updates only that winner with \ac{bc} and critic guidance.
Because the routing is recomputed from the current candidate geometry, ownership of a supported region can shift across candidates over the course of learning.
This gives a one-step actor room to make local improvements that pointwise extraction struggles to capture, while retaining single-pass inference at test time.
On OGBench and D4RL, DROL is competitive with the one-step FQL baseline, improving many OGBench task groups while remaining strong on both AntMaze and Adroit.
Project page: \href{https://muzhancun.github.io/preprints/DROL}{\texttt{https://muzhancun.github.io/preprints/DROL}}.

%% file: section/introduction.tex
\section{Introduction}

Offline \acf{rl} requires improving a policy while staying within the action regions supported by a fixed dataset~\citep{levine2020offline}.
This requirement becomes harder to satisfy on multimodal benchmarks such as OGBench~\citep{ogbench_park2025}, where nearby states can admit several distinct yet reasonable actions.
Generative actors are attractive in this regime because they represent such multimodal behavior better than standard unimodal policies~\citep{diffusion_sohl2015,flow_lipman2023,dql_wang2023,idql_hansenestruch2023,fql_park2025}.

Many strong generative offline \ac{rl} methods rely on iterative sampling during training, inference, or both~\citep{dql_wang2023,idql_hansenestruch2023,espinosa2025scaling,mu2026deflow}.
A one-step actor is attractive because it avoids backpropagating through a long iterative sampler and keeps execution cheap.
One common way to obtain such an actor, exemplified by FQL~\citep{fql_park2025}, is to distill a stronger iterative teacher: for a latent draw $z$, the teacher produces $\tilde a=\mu_\psi(s,z)$, and the student output $f_\theta(s,z)$ is trained to improve Q while staying close to $\tilde a$.
The difficulty is structural: within each update, the same sampled output must serve both roles.
It is pushed toward higher Q while simultaneously being pulled back toward its paired teacher endpoint $\tilde a$.
When these two directions disagree, the loss resolves the conflict by compromising on that same sample.

\begin{figure*}[t!]
    \centering
    \includegraphics[width=\textwidth]{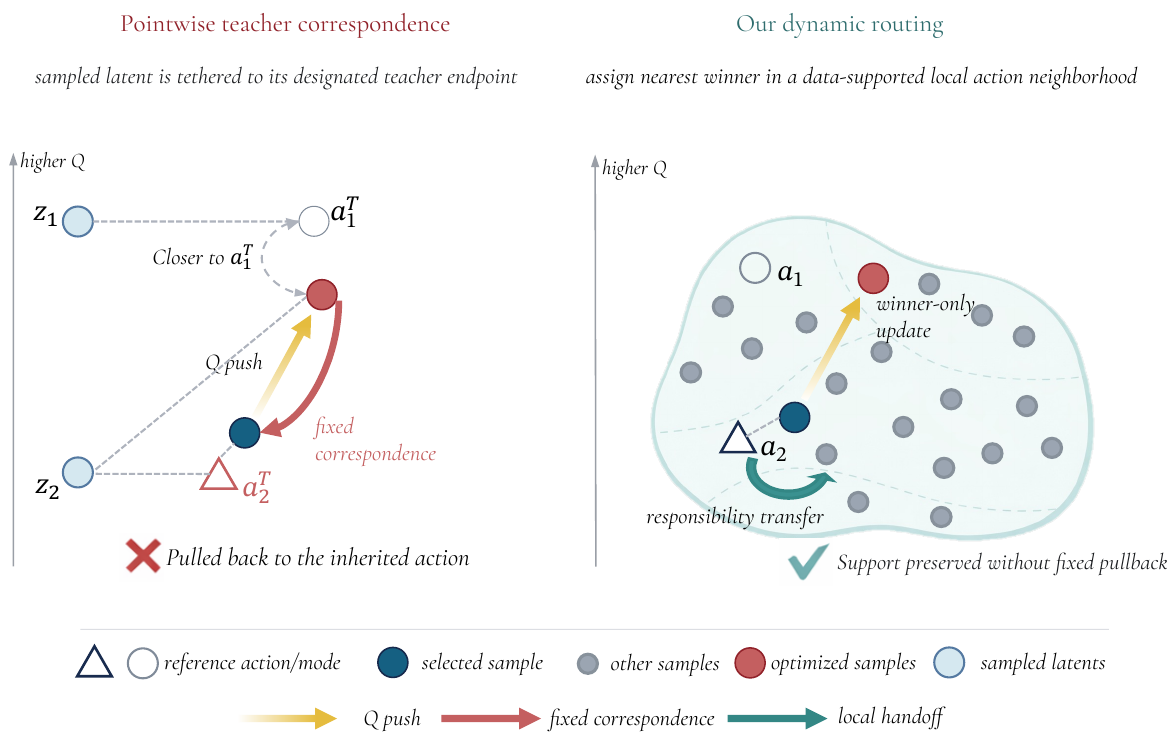}
    \caption{
    Preserve support, not correspondence.
    Left: pointwise extraction assigns both improvement and reconstruction to the same sampled output.
    Right: DROL routes each dataset action to the nearest candidate, so ownership of a supported region can transfer across candidates.
    }
    \label{fig:method_overview}
\end{figure*}

For a multimodal actor, different samples can naturally cover different supported modes.
\emph{The key question is whether that ownership can shift during training.}
If it cannot, a sampled output remains responsible for its original endpoint even when a nearby supported improvement is available.
If it can, that sample may move toward the better action while another takes over the old region.
In FQL-style extraction, ownership stays attached to the same sampled output.
In DROL, it can transfer across candidates.
This tension is illustrated in \Cref{fig:method_overview}.

DROL retains the cheap one-step actor but changes the regularizer.
For each state, it samples $K$ candidate actions, routes each dataset action to its nearest candidate,
$$k^*(s,a)=\arg\min_{k\in[K]} \|\hat a^k-a\|_2^2,$$
and updates only that winner with behavior cloning and Q-improvement.
As shown in \Cref{fig:method_overview}, one candidate can move toward a better local action while another retains ownership of the old supported region.
The key difference from FQL-style extraction is that the behavior-cloning pull can change owners across candidates rather than remaining attached to a single sampled output.
This design allows different latent regions to specialize to different parts of local support while preserving one-step inference at test time.

Our main contributions are:
\begin{enumerate}
\item We identify pointwise latent-to-teacher correspondence as an unnecessary constraint in one-step extraction for multimodal offline \ac{rl}, and argue that local action support is the object worth preserving.
\item We propose DROL, a routed candidate-set actor that samples multiple actions during training, assigns each data action to its nearest candidate, and updates only the routed winner.
\item We analyze the mechanism of routing, including non-collapse, responsibility transfer, and the role of the routing budget $K$, and evaluate the resulting one-step actor on OGBench and D4RL~\citep{d4rl_fu2020} with routing visualizations and $K$ ablations.
\end{enumerate}

%% file: section/preliminaries.tex
\section{Preliminaries and Problem Setup}

We consider an \ac{mdp} $(\mathcal{S}, \mathcal{A}, r, p, \gamma)$ together with a fixed offline dataset $\mathcal{D}=\{(s,a,r,s')\}$.
Given a critic $Q_\phi$, the actor should seek high-value actions while staying close to the state-conditional action regions supported by the data.
At an abstract level, offline \ac{rl} can be written as
\begin{equation}
\begin{aligned}
    \max_{\pi_\theta}\quad & \mathbb{E}_{s \sim \mathcal{D},\, a \sim \pi_\theta(\cdot|s)}[Q_\phi(s,a)] \\
    \textrm{s.t.}\quad & \mathbb{D}\bigl(\pi_\theta(\cdot|s), \pi_\beta(\cdot|s)\bigr) \le \epsilon,
\end{aligned}
\label{eq:constrained_obj}
\end{equation}
where $\pi_\beta$ denotes the implicit behavior distribution and $\mathbb{D}$ is a discrepancy that keeps policy improvement within trustworthy regions.
In the multimodal setting studied here, exact duplicates at a single observation are rare, so we instead search within a small neighborhood of states around $s$ rather than relying on literal repeated supervision at the exact same state.
For this paper, the central question is which object the regularizer is actually trying to preserve.

We train the critic with the usual Bellman regression objective
\begin{equation}
\begin{aligned}
    \mathcal{L}_{Q}(\phi)
    =
    \mathbb{E}_{(s,a,r,s') \sim \mathcal{D}}
    \Bigl[
        \bigl(
        Q_\phi(s,a)
        -
        (r + \gamma Q_{\bar{\phi}}(s', \pi_\theta(s')))
        \bigr)^2
    \Bigr],
\end{aligned}
\label{eq:critic_prelim}
\end{equation}
so the main design question is concentrated in the actor update.

\paragraph{Positioning relative to prior work.}
Among generative offline \ac{rl} methods, FQL~\citep{fql_park2025} is the clearest reference point, since it also trains a one-step actor from a stronger iterative teacher.
DROL departs from that line at the level of the regularizer: the key issue for us is the pointwise latent-conditioned tether used during extraction.
ReFORM~\citep{zhang2026reform} is related in a different way.
It also argues that support should be preserved in offline \ac{rl}, but does so through reflected flow constructions rather than routed candidate sets.
Finally, the routed behavior-cloning update is closest to \ac{imle}~\citep{li2019imle} and \acs{imle} Policy~\citep{rana2025imlepolicy}, which also rely on nearest-sample or set-level matching.
What is new here is using that matching primitive as the actor-side support regularizer, coupled directly to local Q-improvement.
See \cref{app:related_work} for a broader discussion of adjacent generative offline \ac{rl} and policy-learning methods.

\subsection{Pointwise-Correspondence One-Step Extraction}

Many recent generative offline \ac{rl} methods, most notably FQL~\citep{fql_park2025}, use a strong iterative model $\mu_\psi$ as a behavior policy and then train a cheaper one-step actor $f_\theta$ for critic-guided optimization.
The teacher can be instantiated as a flow, diffusion, or other latent-conditioned generator; the precise architecture matters less here than the structure of the extracted actor loss.

Let $z \sim p_0$ be a latent variable shared by the teacher and student, and let $\tilde a = \mu_\psi(s,z)$ be the teacher action associated with $(s,z)$.
A common extraction template is
\begin{equation}
\begin{aligned}
    \mathcal{L}_{\pi}(\theta)
    =
    \mathbb{E}_{s,z}
    \Bigl[
        - Q_\phi(s, f_\theta(s,z))
        +
        \alpha\, \ell\bigl(f_\theta(s,z), \tilde a\bigr)
    \Bigr],
\end{aligned}
\label{eq:fql_style_actor}
\end{equation}
where $\ell$ is a distillation loss such as squared error.
This objective is practical because it avoids differentiating through the teacher's full sampling process.
It also imposes a pointwise correspondence: for each sampled pair $(s,z)$, the same student output $f_\theta(s,z)$ is trained toward the teacher endpoint $\tilde a=\mu_\psi(s,z)$ associated with that latent.
The endpoint may shift with $s$ and with the teacher parameters over training, but within the current gradient the Q term and the distillation term remain attached to that same sampled output.
That output therefore has to absorb the tradeoff on its own: if moving toward a higher-Q action pushes it away from $\tilde a$, the same update pulls it back.

This is the structural foil for DROL.
If the critic prefers a nearby action that remains well supported by the dataset, \cref{eq:fql_style_actor} provides no second sampled output that can take over the old teacher-linked region.
DROL retains the tractable one-step actor setup but replaces samplewise teacher-student alignment with set-level routing over the actor's current candidate actions.

%% file: section/method.tex
\section{Dynamic Routing for Offline RL}
\label{sec:method}

\subsection{Actor as a Candidate Set}

We start from a latent-conditioned actor $f_\theta(s,z)$.
The local support around $s$ is not observed directly, and exact support inclusion is intractable for a continuous generative actor.
DROL therefore routes over a sampled candidate set drawn from the actor itself; the separate local-support object is used only in the conceptual discussion and visualization.

For each state $s$, we sample $K$ latent codes and generate
\begin{equation}
    \hat{a}^k = f_\theta(s, z_k),
    \qquad
    z_k \sim \mathrm{Unif}(B_R^{d_z}),
    \qquad
    k=1,\dots,K,
\end{equation}
where $d_z$ is the latent dimension and $B_R^{d_z}$ is the $d_z$-dimensional Euclidean ball of radius $R$.
Following ReFORM~\citep{zhang2026reform}, we adopt the dimension-aware choice $R=\sqrt{d_a}$ in our experiments, where $d_a$ is the action dimension.
With this choice, the overall latent scale stays on the same order as the action dimension (see \cref{app:proof_scale}).
These samples form the candidate set
\begin{equation}
    A_\theta(s) = \{\hat a^1,\dots,\hat a^K\}.
\end{equation}
A larger candidate set exposes more local directions to the routing rule simultaneously, enabling specialization and subsequent handoff during training.

The choice $R=\sqrt{d_a}$ ties the latent scale to the action geometry rather than fixing a single radius across tasks, while keeping the prior isotropic.
Compared with an unbounded Gaussian prior, $\mathrm{Unif}(B_R^{d_z})$ removes heavy-tail latent excursions and keeps the sampled candidate set compact, which is better aligned with a routing-based local support surrogate.
At test time, the policy uses only a single latent sample,
\begin{equation}
    \pi_\theta(s) = f_\theta(s,z),
    \qquad
    z \sim \mathrm{Unif}(B_R^{d_z}),
\end{equation}
so the multiplicity $K$ is purely a training-time device.
After training, a single latent draw already samples from the reorganized actor, so execution stays within the plain one-step setting.

\subsection{Top-1 Dynamic Routing}

For each offline pair $(s,a)$, DROL routes the dataset action to its current nearest candidate,
\begin{equation}
    k^*(s,a)
    =
    \arg\min_{k \in [K]} \|\hat a^k-a\|_2^2.
\end{equation}
Only the routed winner $\hat a^{k^*(s,a)}$ receives the actor update for that data point.
Routing is recomputed from the current candidate geometry at every gradient step, and gradients are not propagated through the discrete $\arg\min$ itself.

The actor objective is
\begin{equation}
\begin{aligned}
    \mathcal{L}_{\mathrm{actor}}(\theta)
    =
    \mathbb{E}_{(s,a)\sim\mathcal{D},\, z_{1:K}}
    \Bigl[
        \|\hat a^{k^*(s,a)}-a\|_2^2
        -
        \alpha Q_\phi\bigl(s,\hat a^{k^*(s,a)}\bigr)
    \Bigr].
\end{aligned}
\label{eq:drol_actor}
\end{equation}
The first term is winner-only \acl{bc}.
It anchors the currently responsible candidate to the local data neighborhood that routing assigns to it.
The second term performs critic-guided improvement on that same candidate.
Because both terms act on the same routed prototype, \acl{bc} becomes a local support surrogate rather than a global penalty on the full actor.

The critic is trained with the standard Bellman loss
\begin{equation}
\begin{aligned}
    \mathcal{L}_Q(\phi)
    =
    \mathbb{E}_{(s,a,r,s') \sim \mathcal{D}}
    \Bigl[
        \bigl(
        Q_\phi(s,a)
        -
        (r + \gamma Q_{\bar\phi}(s', \pi_\theta(s')))
        \bigr)^2
    \Bigr].
\end{aligned}
\label{eq:drol_critic}
\end{equation}
Routing determines how support is regularized during actor learning, while execution and critic targets remain those of an ordinary one-sample policy.

\begin{algorithm}[t!]
\caption{DROL}
\label{alg:drol}
\small
\begin{algorithmic}[1]
\While{not converged}
\State Sample minibatch $\{(s_i,a_i,r_i,s'_i)\}_{i=1}^B \sim \mathcal{D}$
\State Sample $z_{i,1:K} \sim \mathrm{Unif}(B_R^{d_z})$ and set $\hat a_i^k \gets f_\theta(s_i, z_{i,k})$
\State Route each data action to $k_i^* \gets \arg\min_k \|\hat a_i^k-a_i\|_2^2$
\State Update $\theta$ on $\frac{1}{B}\sum_i \left[\|\hat a_i^{k_i^*}-a_i\|_2^2 - \alpha Q_\phi(s_i,\hat a_i^{k_i^*})\right]$
\State Sample $z'_i \sim \mathrm{Unif}(B_R^{d_z})$ and set $a'_i \gets f_\theta(s'_i, z'_i)$
\State Update $\phi$ on $\frac{1}{B}\sum_i \left[Q_\phi(s_i,a_i)-r_i-\gamma Q_{\bar\phi}(s'_i,a'_i)\right]^2$
\EndWhile
\end{algorithmic}
\end{algorithm}

\subsection{Interpretation}

Top-1 routing induces a state-conditional Euclidean Voronoi partition over the candidate set~\citep{aurenhammer1991voronoi,okabe2000spatial}.
Given $A_\theta(s)=\{\hat a^1,\dots,\hat a^K\}$, the cell of candidate $k$ is
\begin{equation}
    V_k(s)
    :=
    \left\{
        a \in \mathbb{R}^{d_a}
        :
        \|\hat a^k-a\|_2^2 \le \|\hat a^j-a\|_2^2,\ \forall j \in [K]
    \right\}.
\end{equation}
The cells $\{V_k(s)\}_{k=1}^K$ partition the action space up to measure-zero tie boundaries, and the routing rule returns the index of the cell containing the dataset action.
Each dataset action therefore supervises only the candidate whose current cell contains it, and each candidate is responsible only for the portion of the local support that currently falls within its cell, namely $V_k(s)\cap \mathrm{supp}_{\mathrm{local}}(s)$ in the idealized local-support picture.
If one candidate moves toward a better local action, another can inherit responsibility for the old neighborhood as soon as the nearest-neighbor partition changes.
Iterative policies revisit several actions at inference time; DROL revisits several actions only during actor training.
Once latent regions have specialized, execution returns to ordinary one-sample generation.

\begin{figure}[t]
    \centering
    \includegraphics[width=\linewidth]{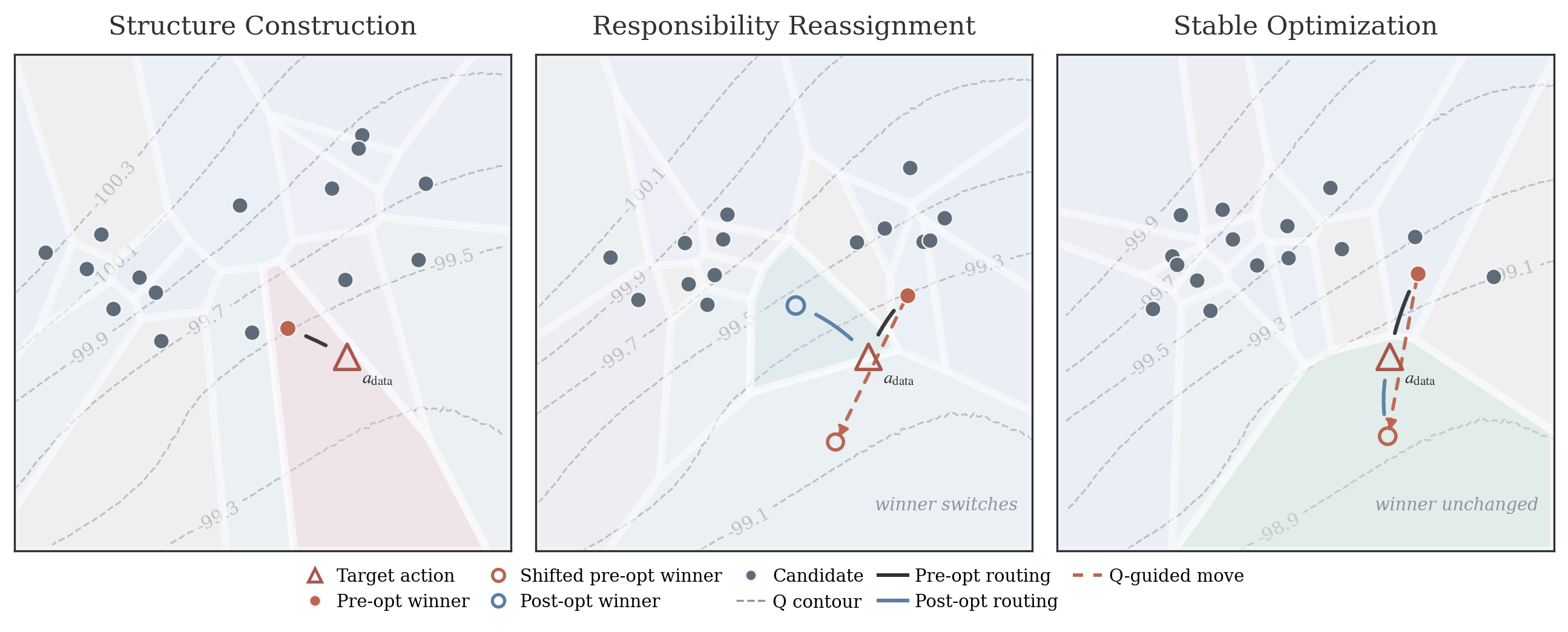}
    \caption{
    Mechanism visualization for DROL.
    The left panel shows structure construction, in which the candidate set spreads rather than collapses.
    The middle and right panels show two common local regimes after that structure has formed: responsibility reassignment and stable optimization.
    These two regimes are not meant to imply a fixed global time order.
    During training, different states can exhibit either behavior, and the same state can revisit both as the actor and critic continue to evolve.
    }
    \label{fig:voronoi_evolution}
\end{figure}

%% file: section/theory.tex
\section{Three Insights Behind DROL}
\label{sec:theory}

This section is organized around the three questions that tend to come up first.
Will winner-only routing collapse the candidates?
Why can one candidate move without immediately destroying support elsewhere?
Why does a larger routing budget $K$ help on some task families?
The statements below are intentionally local: they are meant to explain the actor update rather than prove global \ac{rl} convergence.
Proofs are deferred to \cref{app:proofs}.

\subsection{A Separated-Demand Toy Model Does Not Collapse}

It is helpful to begin with a small dispatch model.
Imagine a fixed local state neighborhood in which action demand concentrates around several separated regions.
Each candidate acts like a rider serving the orders closest to it.
If two riders wait in the same neighborhood while another neighborhood has none, the average service distance increases.
Winner-only routing therefore behaves like local specialization rather than global averaging.

\begin{proposition}[Separated Neighborhoods Avoid Collapse]
Let
\[
\rho
=
\frac{1}{M}\sum_{m=1}^M \mathrm{Unif}(I_m)
\]
be a one-dimensional action distribution supported on disjoint intervals
\[
I_m=[c_m-r,c_m+r],
\qquad
m=1,\dots,M,
\]
with gaps $c_{m+1}-c_m>4r$.
For $K=M$, consider the routed behavior-cloning objective
\[
\mathcal J(\hat a_{1:K})
:=
\int \min_{k \in [K]} |\hat a_k-a|^2 \,\rho(\mathrm{d}a).
\]
Then every global minimizer of $\mathcal J$ places exactly one prototype in each interval $I_m$.
In particular, a collapsed configuration with two prototypes in one interval and none in another cannot be optimal.
\end{proposition}

This toy model addresses the main collapse objection.
Winner-only \acl{bc} does not pull all candidates toward the same point.
When the local support contains separated neighborhoods, the routed reconstruction loss prefers one candidate per neighborhood.
That is precisely the mechanism we want from DROL.
The visualization in \cref{fig:voronoi_evolution} should be read in this spirit: the candidate set spreads to cover a local support cloud before subsequent updates refine it.

\subsection{Routing Replaces a Persistent Tether with a Transferable Anchor}

Consider the gap between a pointwise-correspondence loss and a routed loss for a single supported target $a$ at state $s$.
Under pointwise distillation, $a$ can be read as the designated teacher endpoint for a sampled latent-state pair.
Under routing, it is the local target currently assigned to the winning candidate.
For a single candidate $x$, the pointwise-correspondence objective is
\[
\ell_{\mathrm{fix}}(x;a)
=
\|x-a\|_2^2-\alpha Q_\phi(s,x).
\]
For a routed candidate set $A=(x_1,\dots,x_K)$, define
\[
\ell_{\mathrm{route}}(A;a)
=
\min_{k \in [K]} \|x_k-a\|_2^2
\;-\;
\alpha Q_\phi\bigl(s,x_{k^*(a)}\bigr),
\qquad
k^*(a)=\arg\min_k \|x_k-a\|_2^2.
\]

\begin{proposition}[Persistent Tether versus Routed Anchor]
For the pointwise-correspondence objective,
\[
\nabla_x \ell_{\mathrm{fix}}(x;a)
=
2(x-a)-\alpha \nabla_x Q_\phi(s,x),
\]
so the sample $a$ continues to exert the same pullback term $2(x-a)$ regardless of what other candidates may exist.
For the routed objective, if the winner $k^*(a)$ is unique, then
\[
\nabla_{x_i}\ell_{\mathrm{route}}(A;a)=0
\qquad
\text{for all } i\neq k^*(a),
\]
and
\[
\nabla_{x_{k^*(a)}}\ell_{\mathrm{route}}(A;a)
=
2(x_{k^*(a)}-a)-\alpha \nabla_x Q_\phi(s,x_{k^*(a)}).
\]
Once another candidate becomes the nearest one to $a$, this behavior-cloning contribution transfers to the new winner.
\end{proposition}

\begin{proposition}[Pointwise Correspondence Biases a Local Improvement]
Let $U \subset \mathbb{R}^{d_a}$ be convex, let $Q_\phi(s,\cdot)$ be differentiable and $m$-strongly concave on $U$, and let $x_Q \in U$ be an interior maximizer of $Q_\phi(s,\cdot)$ over $U$.
For a target $a \in U$, define
\[
g(x)
=
\|x-a\|_2^2-\alpha Q_\phi(s,x),
\qquad x \in U.
\]
Then $g$ has a unique minimizer $x_{\mathrm{fix}}$.
Moreover:
\begin{enumerate}
\item $x_Q$ is stationary for $g$ if and only if $a=x_Q$.
\item The minimizer satisfies
\[
\|x_{\mathrm{fix}}-x_Q\|_2
\le
\frac{2}{2+\alpha m}\|a-x_Q\|_2.
\]
\end{enumerate}
\end{proposition}

The partition matters because it determines which candidate receives the pullback from a supported target.
Under pointwise teacher distillation, the same sampled latent-state pair keeps paying to stay near its designated teacher endpoint.
Under routing, that pullback is attached to the current winner of the nearest-neighbor partition, and when the partition changes, the pullback changes owner with it.

The second proposition makes the local consequence explicit.
Even when the inherited target $a$ and the local Q-favored action $x_Q$ lie in the same supported neighborhood, $x_Q$ is not stationary under a pointwise-correspondence loss unless $a=x_Q$.
The optimizer therefore settles at a compromise between the old target and the critic-preferred action.
Routed training does not eliminate the behavior-cloning term, but it allows another candidate to take over the old neighborhood once the partition changes.
This is the local mechanism that enables a one-step actor to follow the critic more freely without dropping support.

\subsection{Larger \texorpdfstring{$K$}{K} Improves Handoff Availability}

The final question is why a larger routing budget helps on some task families.
For a fixed state $s$, suppose the local support is covered by modal neighborhoods $U_1(s),\dots,U_{M_s}(s)$.
Let
\[
p_m(s)
:=
\mathbb P_{z \sim \mathrm{Unif}(B_R^{d_z})}\bigl(f_\theta(s,z) \in U_m(s)\bigr),
\]
so that $p_m(s)$ is the probability that one sampled candidate lands in neighborhood $m$.

\begin{proposition}[Coverage from Larger Candidate Sets]
If $p_m(s)>0$ for all $m=1,\dots,M_s$, then
\[
\mathbb P\Bigl(\forall m,\exists k\in[K],\hat a_k \in U_m(s)\Bigr)
\ge
1-\sum_{m=1}^{M_s}(1-p_m(s))^K.
\]
In particular, letting $p_{\min}(s):=\min_m p_m(s)$,
\[
\mathbb P\Bigl(\forall m,\exists k,\hat a_k \in U_m(s)\Bigr)
\ge
1-M_s e^{-Kp_{\min}(s)}.
\]
\end{proposition}

This is not a coverage-creation statement.
Increasing $K$ does not invent new supported actions; it only makes it less likely that the current candidate set misses neighborhoods that are already reachable.
That matters because responsibility transfer only works if another candidate is available nearby to take over.
This is the geometric role of $K$ in DROL.

%% file: section/experiments.tex
\section{Experiments}
\label{sec:exp}

We use experiments to answer three practical questions on OGBench~\citep{ogbench_park2025}, supported by additional results on D4RL~\citep{d4rl_fu2020}.
First, can routing make a plain one-step actor competitive with heavier iterative policies on multimodal offline RL tasks?
Second, do the candidate sets exhibit the non-collapse, specialization, handoff, and locally stable regimes suggested by \cref{sec:theory}?
Third, how does the routing budget $K$ affect this behavior?

\subsection{Main Results on OGBench and D4RL}

\input{tables/offline}

Table~\ref{tab:ogbench_main} summarizes the main benchmark results.
DROL executes with a single latent draw and no multi-step refinement at test time.
Even under this restriction, the fixed default configuration \texttt{DROL(16)} matches or improves over the one-step FQL baseline on 6 of the 10 OGBench task groups, with especially clear gains on \texttt{antmaze-large}, \texttt{humanoidmaze-medium}, \texttt{antsoccer}, and \texttt{cube-single}.
Selecting $K$ once on a representative default task within each OGBench task family and reusing it across the remaining tasks yields \texttt{DROL}$^*$, which matches or improves over FQL on 9 of the 10 OGBench groups.
The largest gain appears on \texttt{antmaze-giant}, where the family average rises from $9$ for FQL and $3$ for the default configuration to $45$ after tuning only $K$.
The D4RL rows tell a complementary story on a more established benchmark suite: DROL matches FQL on Adroit and remains close on AntMaze, suggesting that the routed one-step actor transfers beyond OGBench rather than fitting a single benchmark family.

\begin{figure*}[!t]
    \centering
    \includegraphics[width=0.95\textwidth]{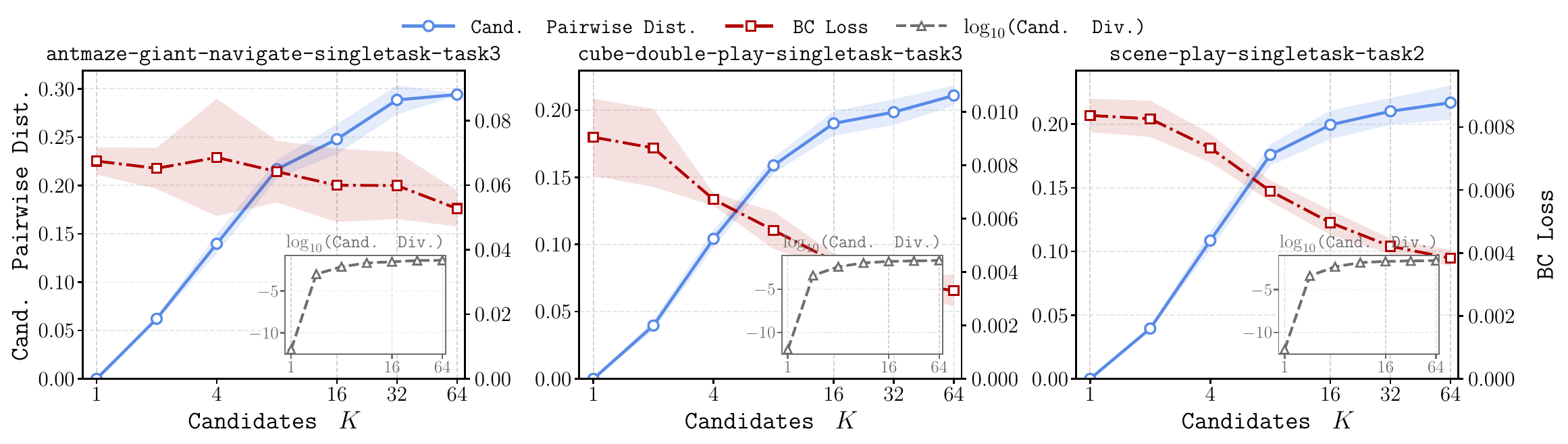}

    \vspace{0.15em}

    \includegraphics[width=0.72\textwidth]{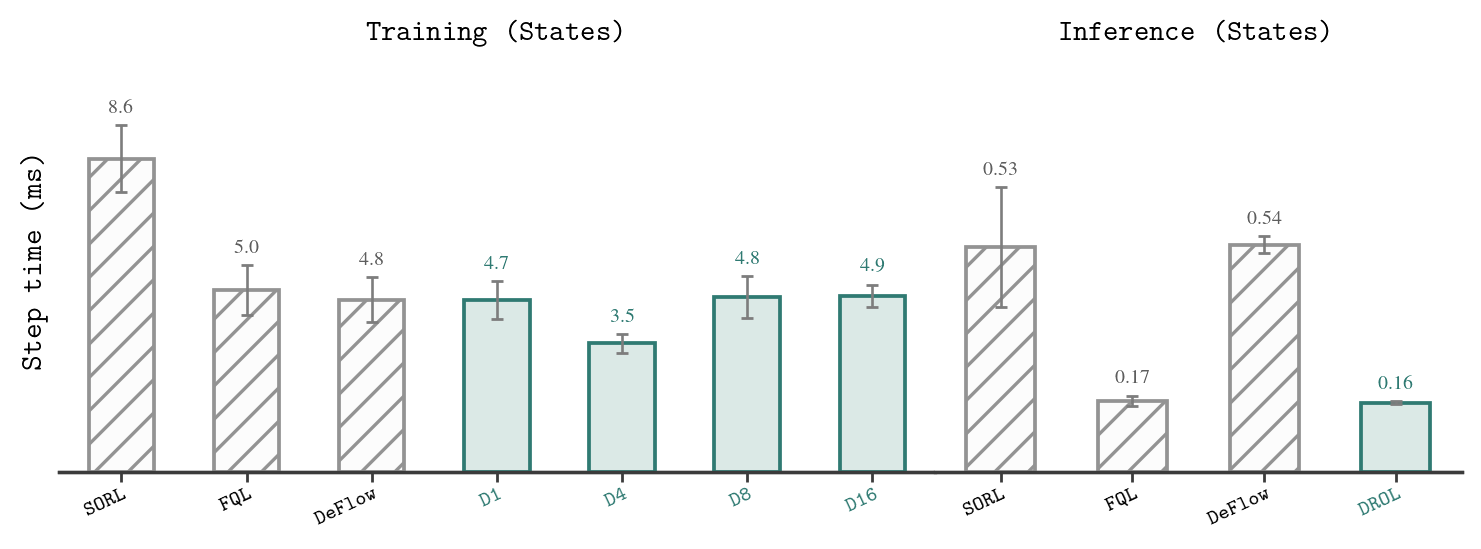}
    \caption{
    Scaling and runtime of routed one-step actors.
    Top: training metrics versus the number of candidates $K$ on three representative environments; \emph{Cand.\ Pairwise Dist.} is the normalized pairwise squared distance, the inset $\log_{10}(\text{Cand.\ Div.})$ shows centroid divergence, and \emph{\acs{bc} Loss} is the routed reconstruction loss.
    Bottom: training and inference step time, where \texttt{D1}/\texttt{D4}/\texttt{D8}/\texttt{D16} denote DROL with routing budget $K \in \{1,4,8,16\}$.
    Larger $K$ increases candidate spread and lowers routed \acs{bc} loss, while inference remains cheap because DROL still executes as a one-step actor.
    }
    \label{fig:k_training_metrics}
\end{figure*}

Taken together, the table supports a narrow claim: once the actor is no longer forced to preserve pointwise teacher correspondence during extraction, a simple one-step policy can already recover a large fraction of the performance that otherwise appears to require heavier inference.
The runtime comparison in \cref{fig:k_training_metrics} tells the same story from the systems side: DROL keeps one-step inference cost essentially at the FQL level while iterative baselines remain substantially slower, and its candidate evaluations are easy to batch across $K$ during training.

\subsection{Mechanism Study: Voronoi Evolution During Training}

\Cref{fig:voronoi_evolution} visualizes the mechanism on a fixed representative state.
Because exact repeated actions at a single observation are rare, the local support object in the discussion is approximated by a dense surrogate at that state, while the figure itself overlays only the current candidate set and its induced Voronoi cells $\{V_k(s)\}_{k=1}^K$ from \cref{sec:method}.
The figure checks three concrete predictions from \cref{sec:theory}: the candidate set should spread rather than collapse, one common regime is responsibility reassignment, and another is stable optimization with an unchanged winner.
The second and third panels should therefore be read as two representative phases of routed training rather than a strict chronology.

\begin{figure*}[!t]
    \centering
    \includegraphics[width=0.91\textwidth]{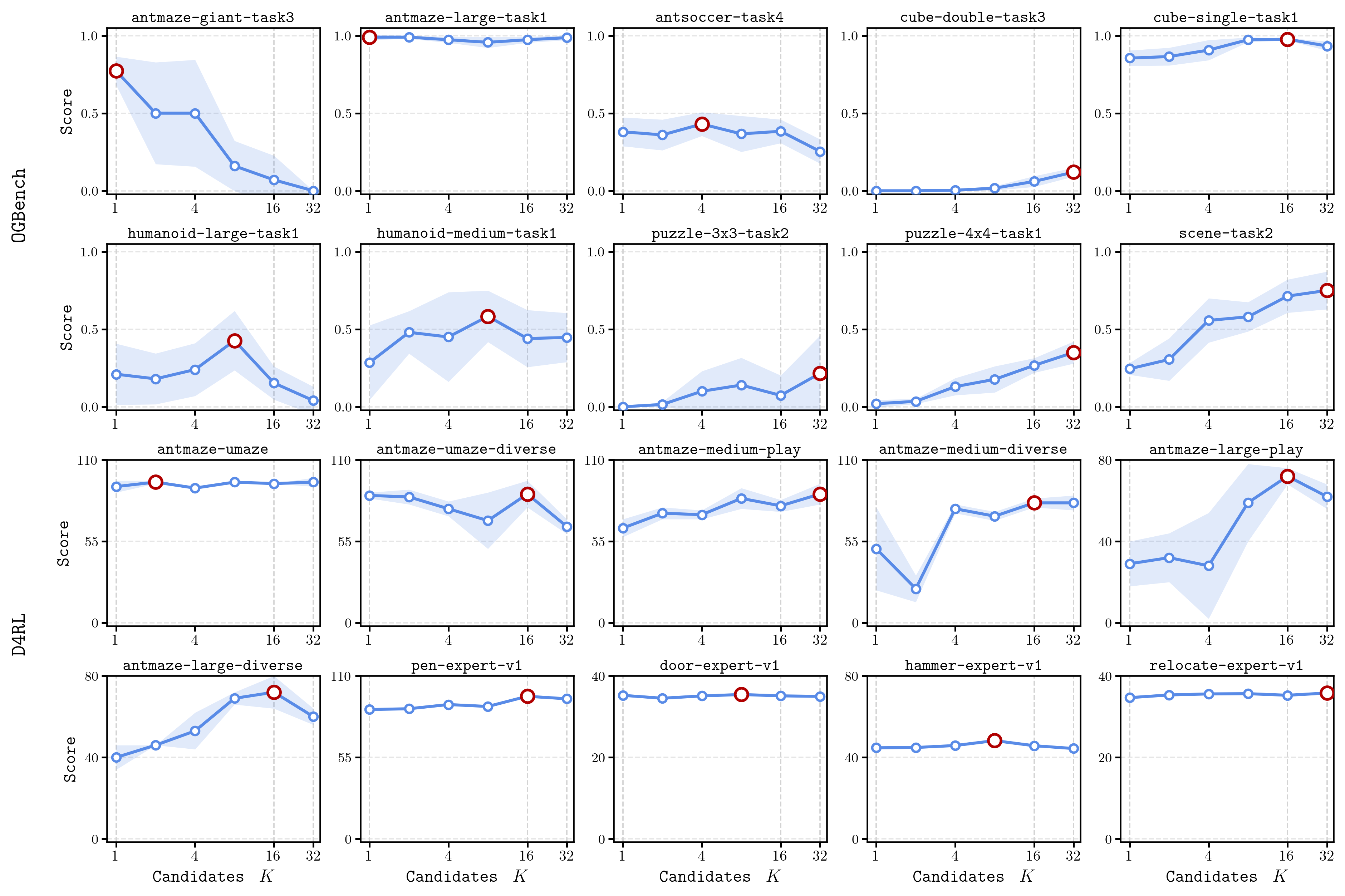}
    \caption{
    Sensitivity to the number of candidates $K$ across OGBench and D4RL.
    Each panel shows one representative sweep over $K \in \{1,2,4,8,16,32\}$; the dashed line marks the default $K=16$ and the red circle marks the best observed $K$.
    }
    \label{fig:k_sensitivity}
\end{figure*}

\subsection{No Collapse and Capacity Scaling with \texorpdfstring{$K$}{K}}

\Cref{fig:k_training_metrics} complements the geometric picture with three training metrics on representative environments.
\emph{Cand.\ Pairwise Dist.} averages $\|\hat a_i-\hat a_j\|_2^2/d_a$ over off-diagonal pairs, the inset $\log_{10}(\text{Cand.\ Div.})$ shows mean centroid divergence $\|\hat a_k-\bar a\|_2^2/d_a$, and the red curve is the routed \acs{bc} loss.

Two trends are consistent across all three environments.
First, the candidate set does not collapse as $K$ grows.
Both the pairwise spread and the centroid divergence increase with $K$, indicating that the additional candidates organize into distinct local actions rather than redundant copies.
Second, the routed \acs{bc} loss decreases with $K$.
This is the scaling one would expect if a larger candidate set provides a finer local quantization of the supported action neighborhood.
Taken together with \cref{fig:voronoi_evolution}, these curves support the picture in \cref{sec:theory}: routing uses larger candidate sets to build more local structure, not to replicate the same action multiple times.

\subsection{Sensitivity to the Number of Candidates \texorpdfstring{$K$}{K}}

\Cref{fig:k_sensitivity} shows that $K$ controls routing capacity rather than acting as a monotone scale parameter.
Large or highly multimodal navigation families benefit most from larger candidate sets, while other families saturate earlier or prefer intermediate values of $K$.
In practice, fixing $K=16$ already makes the workable range of $\alpha$ clear, so we tune $\alpha$ first and sweep $K$ only when additional budget is available.

%% file: tables/offline.tex
\begin{table*}[t!]
\vspace{-5pt}
\caption{
\footnotesize
\textbf{Offline RL results on OGBench and D4RL.}
\texttt{DROL(16)} uses the fixed default $K=16$ throughout. \texttt{DROL}$^*$ uses a benchmark-specific $K$-selection protocol; see Appendix~A for the exact OGBench and D4RL settings.
Bold entries are at least $0.95\times$ the best score in their row, and underlined entries are one-step methods that are at least $0.95\times$ the best score.
See \Cref{table:offline_full} for the full task-level version and \Cref{fig:k_sensitivity} for representative $K$-sweep trends.
}
\label{tab:ogbench_main}
\centering
\vspace{5pt}
\setlength{\tabcolsep}{2.5pt}
\scalebox{0.62}{
\begin{threeparttable}
\begin{tabular}{lccccccccccc}
\toprule
\multicolumn{1}{c}{} & \multicolumn{3}{c}{\texttt{Gaussian Policies}} & \multicolumn{5}{c}{\texttt{Multi-step / Iterative Policies}$^\dagger$} & \multicolumn{3}{c}{\texttt{One-step Methods}} \\
\cmidrule(lr){2-4} \cmidrule(lr){5-9} \cmidrule(lr){10-12}
\texttt{Task Category} & \texttt{BC} & \texttt{IQL} & \texttt{ReBRAC} & \texttt{FAWAC} & \texttt{FBRAC} & \texttt{IFQL} & \texttt{DeFlow} & \texttt{SORL} & \texttt{\color{myblue}FQL} & \texttt{\color{mygreen}DROL(16)} & \texttt{\color{myteal}DROL}$^*$ \\
\midrule
\texttt{OGBench antmaze-large-singletask ($\mathbf{5}$ tasks)} &    $11$ {\tiny $\pm 1$} &    $53$ {\tiny $\pm 3$} &    $81$ {\tiny $\pm 5$} &    $6$ {\tiny $\pm 1$} &    $60$ {\tiny $\pm 6$} &    $28$ {\tiny $\pm 5$} &    $81$ {\tiny $\pm 3$} &    $\mathbf{89}$ {\tiny $\pm 2$} &    $79$ {\tiny $\pm 3$} &    $\underline{\mathbf{90}}$ {\tiny $\pm 2$} &    $\underline{\mathbf{89}}$ {\tiny $\pm 6$} \\
\texttt{OGBench antmaze-giant-singletask ($\mathbf{5}$ tasks)} &    $0$ {\tiny $\pm 0$} &    $4$ {\tiny $\pm 1$} &    $26$ {\tiny $\pm 8$} &    $0$ {\tiny $\pm 0$} &    $4$ {\tiny $\pm 4$} &    $3$ {\tiny $\pm 2$} &    $12$ {\tiny $\pm 5$} &    $9$ {\tiny $\pm 6$} &    $9$ {\tiny $\pm 6$} &    $3$ {\tiny $\pm 3$} &    $\underline{\mathbf{45}}$ {\tiny $\pm 24$} \\
\texttt{OGBench humanoidmaze-medium-singletask ($\mathbf{5}$ tasks)} &    $2$ {\tiny $\pm 1$} &    $33$ {\tiny $\pm 2$} &    $22$ {\tiny $\pm 8$} &    $19$ {\tiny $\pm 1$} &    $38$ {\tiny $\pm 5$} &    $60$ {\tiny $\pm 14$} &    $48$ {\tiny $\pm 4$} &    $64$ {\tiny $\pm 4$} &    $58$ {\tiny $\pm 5$} &    $\underline{\mathbf{65}}$ {\tiny $\pm 9$} &    $\underline{\mathbf{68}}$ {\tiny $\pm 7$} \\
\texttt{OGBench humanoidmaze-large-singletask ($\mathbf{5}$ tasks)} &    $1$ {\tiny $\pm 0$} &    $2$ {\tiny $\pm 1$} &    $2$ {\tiny $\pm 1$} &    $0$ {\tiny $\pm 0$} &    $2$ {\tiny $\pm 0$} &    $\mathbf{11}$ {\tiny $\pm 2$} &    $5$ {\tiny $\pm 2$} &    $5$ {\tiny $\pm 2$} &    $4$ {\tiny $\pm 2$} &    $7$ {\tiny $\pm 4$} &    $\underline{\mathbf{11}}$ {\tiny $\pm 6$} \\
\texttt{OGBench antsoccer-arena-singletask ($\mathbf{5}$ tasks)} &    $1$ {\tiny $\pm 0$} &    $8$ {\tiny $\pm 2$} &    $0$ {\tiny $\pm 0$} &    $12$ {\tiny $\pm 0$} &    $16$ {\tiny $\pm 1$} &    $33$ {\tiny $\pm 6$} &    $\mathbf{67}$ {\tiny $\pm 3$} &    $\mathbf{69}$ {\tiny $\pm 2$} &    $60$ {\tiny $\pm 2$} &    $62$ {\tiny $\pm 2$} &    $65$ {\tiny $\pm 7$} \\
\texttt{OGBench cube-single-singletask ($\mathbf{5}$ tasks)} &    $5$ {\tiny $\pm 1$} &    $83$ {\tiny $\pm 3$} &    $91$ {\tiny $\pm 2$} &    $81$ {\tiny $\pm 4$} &    $79$ {\tiny $\pm 7$} &    $79$ {\tiny $\pm 2$} &    $\mathbf{96}$ {\tiny $\pm 2$} &    $\mathbf{97}$ {\tiny $\pm 1$} &    $\underline{\mathbf{96}}$ {\tiny $\pm 1$} &    $\underline{\mathbf{97}}$ {\tiny $\pm 2$} &    $\underline{\mathbf{98}}$ {\tiny $\pm 2$} \\
\texttt{OGBench cube-double-singletask ($\mathbf{5}$ tasks)} &    $2$ {\tiny $\pm 1$} &    $7$ {\tiny $\pm 1$} &    $12$ {\tiny $\pm 1$} &    $5$ {\tiny $\pm 2$} &    $15$ {\tiny $\pm 3$} &    $14$ {\tiny $\pm 3$} &    $\mathbf{40}$ {\tiny $\pm 5$} &    $25$ {\tiny $\pm 3$} &    $29$ {\tiny $\pm 2$} &    $20$ {\tiny $\pm 3$} &    $26$ {\tiny $\pm 7$} \\
\texttt{OGBench scene-singletask ($\mathbf{5}$ tasks)} &    $5$ {\tiny $\pm 1$} &    $28$ {\tiny $\pm 1$} &    $41$ {\tiny $\pm 3$} &    $30$ {\tiny $\pm 3$} &    $45$ {\tiny $\pm 5$} &    $30$ {\tiny $\pm 3$} &    $51$ {\tiny $\pm 3$} &    $\mathbf{57}$ {\tiny $\pm 2$} &    $\underline{\mathbf{56}}$ {\tiny $\pm 2$} &    $54$ {\tiny $\pm 3$} &    $\underline{\mathbf{58}}$ {\tiny $\pm 3$} \\
\texttt{OGBench puzzle-3x3-singletask ($\mathbf{5}$ tasks)} &    $2$ {\tiny $\pm 0$} &    $9$ {\tiny $\pm 1$} &    $21$ {\tiny $\pm 1$} &    $6$ {\tiny $\pm 2$} &    $14$ {\tiny $\pm 4$} &    $19$ {\tiny $\pm 1$} &    $\mathbf{43}$ {\tiny $\pm 4$} &    $-$ &    $30$ {\tiny $\pm 1$} &    $\underline{\mathbf{41}}$ {\tiny $\pm 9$} &    $\underline{\mathbf{41}}$ {\tiny $\pm 9$} \\
\texttt{OGBench puzzle-4x4-singletask ($\mathbf{5}$ tasks)} &    $0$ {\tiny $\pm 0$} &    $7$ {\tiny $\pm 1$} &    $14$ {\tiny $\pm 1$} &    $1$ {\tiny $\pm 0$} &    $13$ {\tiny $\pm 1$} &    $\mathbf{25}$ {\tiny $\pm 5$} &    $11$ {\tiny $\pm 2$} &    $-$ &    $17$ {\tiny $\pm 2$} &    $16$ {\tiny $\pm 2$} &    $\underline{\mathbf{25}}$ {\tiny $\pm 6$} \\
\texttt{D4RL antmaze ($\mathbf{6}$ tasks)} &    $17$ &    $57$ &    $78$ &    $44$ {\tiny $\pm 3$} &    $64$ {\tiny $\pm 7$} &    $65$ {\tiny $\pm 7$} &    $\mathbf{83}$ {\tiny $\pm 2$} &    $-$ &    $\underline{\mathbf{84}}$ {\tiny $\pm 3$} &    $78$ {\tiny $\pm 8$} &    $\underline{\mathbf{81}}$ {\tiny $\pm 6$} \\
\texttt{D4RL adroit ($\mathbf{12}$ tasks)} &    $48$ &    $53$ &    $\mathbf{59}$ &    $48$ {\tiny $\pm 1$} &    $50$ {\tiny $\pm 2$} &    $52$ {\tiny $\pm 1$} &    $52$ {\tiny $\pm 1$} &    $-$ &    $52$ {\tiny $\pm 1$} &    $51$ {\tiny $\pm 3$} &    $52$ {\tiny $\pm 3$} \\
\bottomrule
\end{tabular}
\begin{tablenotes}
\item[1] \texttt{DROL}$^*$ uses the benchmark-specific $K$-selection protocol described in Appendix~A; all other training settings are unchanged from \texttt{DROL(16)}.
\item[2] The group headed by $^\dagger$ denotes multi-step or iterative policies.
\end{tablenotes}
\end{threeparttable}
\hspace{0pt}}
\end{table*}

%% file: section/conclusion.tex
\section{Conclusion and Limitations}

DROL studies one-step actor learning in multimodal offline \ac{rl} from a geometric perspective.
The actor produces a small candidate set, routing selects the candidate closest to the supported target, and responsibility can transfer as optimization progresses.
This keeps local support neighborhoods intact while still allowing the actor to follow value improvements, and it is already sufficient for a plain one-sample policy to recover a substantial fraction of the performance typically associated with heavier iterative inference, while keeping deployment cost close to that of standard one-step baselines.
In simple terms, routed training continually reconsiders which sampled action should represent the local data neighborhood, and that answer is allowed to change as better actions emerge.

Our theory is local and mechanism-oriented rather than a full convergence theory for offline \ac{rl}, and the method still relies on a fixed routing budget $K$, Euclidean matching, and additional training-time candidate evaluations whose cost grows with $K$.
These limitations point to several natural next steps: learned routing metrics, adaptive candidate budgets, softer responsibility sharing, and region-level regularization beyond pointwise targets.
It is also worth investigating whether geometric constraints can better balance behavior cloning against Q-improvement, and whether best-of-$N$ sampling can further improve the resulting one-step policy.

%% file: section/references.tex
\bibliographystyle{plainnat}

\input{main.bbl}

%% file: section/appendix_implementation.tex
\section{Implementation Details}

We use the same offline training and evaluation pipeline as the one-step FQL baseline.
For each state $s$, DROL samples $K$ latent noises uniformly from the radius-$\sqrt{d_a}$ ball, maps them through a one-step vector field, and obtains $K$ candidate actions.
The dataset action is routed to its nearest candidate under squared Euclidean distance, and both the \acs{bc} term and the Q-improvement term are applied only to that winning candidate.
All reported results use one-step inference.
We do not use best-of-$N$ selection in training or evaluation.

\begin{table}[!hbp]
    \centering
    \caption{Shared architecture and optimization settings used by DROL unless overridden by benchmark-specific flags.}
    \label{tab:impl_defaults}
    \small
    \begin{tabular}{ll}
        \toprule
        Setting & Value \\
        \midrule
        Optimizer & Adam, learning rate $3 \times 10^{-4}$ \\
        Batch size & $256$ \\
        Actor / critic & $4$-layer MLPs with hidden widths $(512,512,512,512)$ \\
        Critic ensemble & $2$ \\
        Target update rate $\tau$ & $0.005$ \\
        Default discount $\gamma$ & $0.99$ \\
        Default routing budget $K$ & $16$ \\
        Offline steps & $10^6$ on OGBench, $5 \times 10^5$ on D4RL \\
        Evaluation & every $100{,}000$ updates, $50$ episodes \\
        \bottomrule
    \end{tabular}
\end{table}

For OGBench, \texttt{DROL(16)} keeps the default $K=16$ in every family.
For \texttt{DROL}$^*$, we choose one family-level routing budget $K^\star$ on the representative task marked by \texttt{(*)} in \Cref{table:offline_full} and reuse it for the other tasks in that family without changing any other hyperparameters.
For D4RL, all runs use $500$K offline updates.
The \texttt{DROL}$^*$ protocol is benchmark-specific.
On D4RL Adroit, we sweep $K$ with the natural task families in mind (for example, \texttt{pen-*}, \texttt{door-*}, \texttt{hammer-*}, and \texttt{relocate-*}) and report the resulting choices in \Cref{tab:impl_ogbench}.
On D4RL AntMaze, we select $K$ per task rather than once per family because the transfer between tasks is noticeably weaker.
The table below summarizes the benchmark-level $(\alpha, K)$ settings.
The right column is only used when \texttt{DROL}$^*$ differs from \texttt{DROL(16)}.

\begin{table}[!hbp]
    \centering
    \caption{Benchmark-level hyperparameters. Each pair is $(\alpha, K)$.}
    \label{tab:impl_ogbench}
    \small
    \begin{tabular}{lccc}
        \toprule
        Dataset / family & \texttt{DROL(16)} & \texttt{DROL}$^*$ if different & Extra flags \\
        \midrule
        \multicolumn{4}{l}{\textbf{OGBench}} \\
        \texttt{antmaze-giant} & $(0.03, 16)$ & $(0.03, 1)$ & \texttt{discount=0.995}, \texttt{q\_agg=min} \\
        \texttt{antmaze-large} & $(0.03, 16)$ & -- & \texttt{discount=0.995}, \texttt{q\_agg=min} \\
        \texttt{antsoccer-arena} & $(0.1, 16)$ & $(0.1, 4)$ & \texttt{discount=0.995} \\
        \texttt{humanoidmaze-medium} & $(0.3, 16)$ & -- & \texttt{discount=0.995} \\
        \texttt{humanoidmaze-large} & $(0.3, 16)$ & $(0.3, 4)$ & \texttt{discount=0.995} \\
        \texttt{cube-single} & $(10.0, 16)$ & -- & -- \\
        \texttt{cube-double} & $(3.0, 16)$ & $(3.0, 64)$ & -- \\
        \texttt{scene} & $(3.0, 16)$ & $(3.0, 64)$ & -- \\
        \texttt{puzzle-3x3} & $(0.3, 16)$ & $(0.3, 32)$ & -- \\
        \texttt{puzzle-4x4} & $(10.0, 16)$ & $(10.0, 64)$ & -- \\
        \midrule
        \multicolumn{4}{l}{\textbf{D4RL}} \\
        \texttt{antmaze-umaze} & $(0.3, 16)$ & -- & -- \\
        \texttt{antmaze-umaze-diverse} & $(0.1, 16)$ & -- & -- \\
        \texttt{antmaze-medium-play} & $(0.1, 16)$ & -- & -- \\
        \texttt{antmaze-medium-diverse} & $(0.1, 16)$ & -- & -- \\
        \texttt{antmaze-large-play} & $(0.03, 16)$ & -- & -- \\
        \texttt{antmaze-large-diverse} & $(0.03, 16)$ & -- & -- \\
        \texttt{pen-human} & $(1.0, 16)$ & $(1.0, 4)$ & \texttt{q\_agg=min} \\
        \texttt{pen-cloned / pen-expert} & $(1.0, 16)$ & -- & \texttt{q\_agg=min} \\
        \texttt{door} & $(10.0, 16)$ & $(10.0, 8)$ & \texttt{q\_agg=min} \\
        \texttt{hammer} & $(3.0, 16)$ & -- & \texttt{q\_agg=min} \\
        \texttt{relocate} & $(10.0, 16)$ & $(10.0, 32)$ & \texttt{q\_agg=min} \\
        \bottomrule
    \end{tabular}
\end{table}

%% file: section/related_work.tex
\section{Related Work}
\label{app:related_work}

Classical offline RL methods such as CQL~\citep{cql_kumar2020}, TD3+BC~\citep{td3bc_fujimoto2021}, and IQL~\citep{iql_kostrikov2022} control distribution shift mainly through critic-side pessimism or simple behavior regularization.
As multimodal benchmarks such as OGBench have made actor expressiveness more important, recent work has increasingly turned to diffusion and flow policies, including Diffusion-QL~\citep{dql_wang2023}, IDQL~\citep{idql_hansenestruch2023}, and flow-matching-based generators built on \citet{flow_lipman2023}.

DROL is most closely related to generative offline RL methods that focus on policy extraction once the critic is fixed or largely standard.
FQL~\citep{fql_park2025} is the clearest reference point: it learns an expressive flow teacher and distills it into a one-step actor for critic-guided optimization.
MeanFlow-QL~\citep{wang2025one}, SORL~\citep{espinosa2025scaling}, FAC~\citep{chae2026fac}, and DeFlow~\citep{mu2026deflow} pursue adjacent efficiency or refinement strategies.
Our departure from this family is not the specific teacher architecture, but the object being regularized during extraction: teacher-student objectives of this type typically preserve a pointwise latent-conditioned pairing between teacher endpoints and actor outputs, whereas DROL regularizes the actor's current candidate set at the level of local supported neighborhoods.

Another closely related direction emphasizes support-aware flow design.
ReFORM~\citep{zhang2026reform} is especially relevant because it also argues that support, rather than generic distributional closeness, should be the operative constraint in offline RL, and it uses bounded and reflected flow constructions to preserve support by design.
DROL is complementary in spirit but different in mechanism: rather than constraining policy improvement through a reflected flow transformation, it uses dynamic routing over a one-step candidate set and lets responsibility move between candidates as the local partition changes.

At the policy-learning level, the routed behavior-cloning term is most closely related to \acs{imle}~\citep{li2019imle} and to its policy instantiation in \acs{imle} Policy~\citep{rana2025imlepolicy}.
These methods train a conditional generator through nearest-sample or set-level matching rather than pointwise sample-to-target correspondence, which is why they are the right comparison point for DROL's winner-only \acs{bc} objective.
The novelty here is not the matching primitive itself, but its use as the support-aware regularizer coupled directly to local Q-improvement on the routed winner.
DROL also differs from IBC~\citep{florence2022ibc}, which models the policy implicitly with an energy-based objective rather than a direct latent-conditioned generator with hard routing during training.

%% file: section/appendix_proofs.tex
\section{Proof Details}
\label{app:proofs}

This appendix records proofs for the propositions in \cref{sec:theory}.
The goal is to make the mechanism-level claims explicit.
We do not attempt a full convergence theory for offline RL.

\subsection{Separated Neighborhoods Avoid Collapse}

Write the routed behavior-cloning objective as
\[
\mathcal J(\hat a_{1:K})
=
\int \min_{k \in [K]} |\hat a_k-a|^2 \,\rho(\mathrm{d}a)
\]
for the separated interval mixture in the proposition.
For $q\ge 1$, let $D_q$ denote the minimum distortion contributed by one interval when $q$ prototypes are assigned to that interval alone.
After translating the interval to $[-r,r]$, any Voronoi partition into cell lengths $\ell_1,\dots,\ell_q$ with $\sum_i \ell_i=2r$ contributes
\[
\frac{1}{2r}\sum_{i=1}^q \frac{\ell_i^3}{12}.
\]
By Jensen's inequality,
\[
\frac{1}{2r}\sum_{i=1}^q \frac{\ell_i^3}{12}
\ge
\frac{q}{24r}\Bigl(\frac{2r}{q}\Bigr)^3
=
\frac{r^2}{3q^2},
\]
with equality for equal cells and prototypes at the cell midpoints. Hence
\[
D_q=\frac{r^2}{3q^2},
\qquad
D_1=\frac{r^2}{3},
\qquad
D_2=\frac{r^2}{12}.
\]

Now consider a global minimizer and suppose, for contradiction, that some interval $I_j$ contains no prototype.
In an optimal one-dimensional quantizer, every active prototype can be placed at the mean of the support mass assigned to its Voronoi cell, so with $K=M$ an empty interval forces some other interval $I_\ell$ to contain at least $q\ge 2$ prototypes.

For any $x\notin I_j$ and $a\sim\mathrm{Unif}(I_j)$,
\[
\mathbb E|x-a|^2
=
(x-c_j)^2+\frac{r^2}{3}.
\]
Since $x\notin I_j$, we have $|x-c_j|\ge r$, and therefore
\[
\mathbb E|x-a|^2 \ge \frac{4r^2}{3}.
\]
So the contribution of $I_j$ to $\mathcal J$ is at least $\frac{4r^2}{3M}$.

Move one prototype out of $I_\ell$ and place it at the midpoint $c_j$ of $I_j$.
Reoptimizing the remaining $q-1$ prototypes on $I_\ell$, the new contribution of $I_j$ is exactly $\frac{1}{M}D_1=\frac{r^2}{3M}$, while the increase on $I_\ell$ is at most
\[
\frac{1}{M}(D_{q-1}-D_q)
\le
\frac{1}{M}(D_1-D_2)
=
\frac{r^2}{4M}.
\]
The total change in the objective is therefore at most
\[
\frac{r^2}{3M}+\frac{r^2}{4M}-\frac{4r^2}{3M}
=
-\frac{3r^2}{4M}
<0,
\]
which contradicts global optimality.

Hence every interval contains at least one prototype.
Since there are exactly $M$ intervals and $K=M$ prototypes, every interval contains exactly one prototype.

\subsection{Persistent Tether versus Routed Anchor}

For the pointwise-correspondence loss,
\[
\ell_{\mathrm{fix}}(x;a)=\|x-a\|_2^2-\alpha Q_\phi(s,x),
\]
the gradient is obtained directly by differentiation:
\[
\nabla_x \ell_{\mathrm{fix}}(x;a)=2(x-a)-\alpha \nabla_x Q_\phi(s,x).
\]
The pullback term $2(x-a)$ is always attached to the same variable $x$.

For the routed loss,
\[
\ell_{\mathrm{route}}(A;a)
=
\min_{k \in [K]} \|x_k-a\|_2^2-\alpha Q_\phi(s,x_{k^*(a)}),
\]
fix a point where the winner $k^*(a)$ is unique and write $k^*=k^*(a)$.
Define the routing margin
\[
\delta
:=
\min_{j\neq k^*}\Bigl(\|x_j-a\|_2^2-\|x_{k^*}-a\|_2^2\Bigr)>0.
\]
By continuity of squared distance, there is a neighborhood of $A$ in which this strict ordering is preserved, so the winner remains $k^*$ throughout that neighborhood.
On that neighborhood,
\[
\ell_{\mathrm{route}}(A;a)
=
\|x_{k^*}-a\|_2^2-\alpha Q_\phi(s,x_{k^*}).
\]
Hence the loss is locally independent of every non-winner $x_i$ with $i\neq k^*$, which gives
\[
\nabla_{x_i}\ell_{\mathrm{route}}(A;a)=0,
\qquad
i\neq k^*,
\]
and
\[
\nabla_{x_{k^*}}\ell_{\mathrm{route}}(A;a)
=
2(x_{k^*}-a)-\alpha \nabla_x Q_\phi(s,x_{k^*}).
\]
When the nearest candidate changes, the winner index changes as well, so the same behavior-cloning term is attached to a different prototype.

\subsection{Fixed Correspondence Biases a Local Improvement}

Let
\[
g(x)=\|x-a\|_2^2-\alpha Q_\phi(s,x)
\]
on the convex set $U$.
Since $Q_\phi(s,\cdot)$ is $m$-strongly concave, the function $-Q_\phi(s,\cdot)$ is $m$-strongly convex.
The quadratic term $\|x-a\|_2^2$ is $2$-strongly convex, so $g$ is $(2+\alpha m)$-strongly convex on $U$.
Strong convexity on the convex set $U$ implies that $g$ has a unique minimizer $x_{\mathrm{fix}}$.

Because $x_Q$ is an interior maximizer of $Q_\phi(s,\cdot)$, we have $\nabla_x Q_\phi(s,x_Q)=0$.
Therefore
\[
\nabla g(x_Q)=2(x_Q-a).
\]
This vanishes if and only if $a=x_Q$, which proves the first claim.

For the second claim, use strong monotonicity of $\nabla g$:
\[
\langle \nabla g(x_{\mathrm{fix}})-\nabla g(x_Q),\, x_{\mathrm{fix}}-x_Q \rangle
\ge
(2+\alpha m)\|x_{\mathrm{fix}}-x_Q\|_2^2.
\]
Since $\nabla g(x_{\mathrm{fix}})=0$ and $\nabla g(x_Q)=2(x_Q-a)$,
\[
-2\langle x_Q-a,\, x_{\mathrm{fix}}-x_Q \rangle
\ge
(2+\alpha m)\|x_{\mathrm{fix}}-x_Q\|_2^2.
\]
Applying Cauchy--Schwarz and canceling one factor of $\|x_{\mathrm{fix}}-x_Q\|_2$ yields
\[
\|x_{\mathrm{fix}}-x_Q\|_2
\le
\frac{2}{2+\alpha m}\|a-x_Q\|_2.
\]
Thus the optimizer stays biased toward the designated teacher endpoint even when $x_Q$ lies in the same local support neighborhood.

\subsection{Coverage from Larger Candidate Sets}

Fix $s$ and a modal neighborhood $U_m(s)$.
For each $m$, define the miss event
\[
E_m
:=
\{\hat a_k \notin U_m(s)\text{ for all }k\in[K]\}.
\]
Because the $K$ latent samples are independent and each single sample lands in $U_m(s)$ with probability $p_m(s)$,
\[
\mathbb P(E_m)=(1-p_m(s))^K.
\]
The event that some neighborhood is missed is $\bigcup_{m=1}^{M_s}E_m$, so a union bound gives
\[
\mathbb P\Bigl(\exists m \text{ such that all } K \text{ candidates miss } U_m(s)\Bigr)
\le
\sum_{m=1}^{M_s}(1-p_m(s))^K.
\]
Taking complements yields the first inequality in the proposition.
For the exponential form, use $(1-p)^K \le e^{-Kp}$ together with $p_m(s)\ge p_{\min}(s)$:
\[
\sum_{m=1}^{M_s}(1-p_m(s))^K
\le
\sum_{m=1}^{M_s}e^{-Kp_m(s)}
\le
M_s e^{-Kp_{\min}(s)}.
\]
Substituting this into the previous display gives the stated bound.

\subsection{Scale of the Bounded Latent Prior}
\label{app:proof_scale}

Let $z \sim \mathrm{Unif}(B_R^{d_z})$.
By rotational symmetry, the radius $r=\|z\|_2$ has density
\[
f(r)=\frac{d_z\, r^{d_z-1}}{R^{d_z}},
\qquad
0 \le r \le R.
\]
Therefore
\[
\mathbb E\|z\|_2^2
=
\int_0^R r^2 f(r)\,\mathrm{d}r
=
\frac{d_z}{R^{d_z}}\int_0^R r^{d_z+1}\,\mathrm{d}r
=
\frac{d_z}{d_z+2}R^2.
\]
With the ReFORM choice $R=\sqrt{d_a}$, this becomes
\[
\mathbb E\|z\|_2^2
=
\frac{d_z}{d_z+2}d_a,
\]
so the overall latent scale is of the same order as the action dimension.

%% file: section/appendix_results.tex
\section{Additional OGBench and D4RL Results}

This appendix reports the full task-level benchmark breakdown corresponding to the grouped OGBench and D4RL summaries in the main text.

\input{tables/offline_full}

%% file: tables/offline_full.tex
\clearpage

\thispagestyle{empty}
\begin{table*}[t!]
\vspace{-30pt}
\caption{
\footnotesize
\textbf{Full task-level benchmark layout on OGBench and D4RL.}
\texttt{DROL(16)} uses the fixed default $K=16$. \texttt{DROL}$^*$ uses the benchmark-specific $K$-selection protocol described in Appendix~A.
Bold entries are at least $0.95\times$ the best score in their row, and underlined entries are one-step methods that are at least $0.95\times$ the best score.
}
\label{table:offline_full}
\centering
\vspace{5pt}
\setlength{\tabcolsep}{2.5pt}
\scalebox{0.62}{
\begin{threeparttable}
\begin{tabular}{lccccccccccc}
\toprule
\multicolumn{1}{c}{} & \multicolumn{3}{c}{\texttt{Gaussian Policies}} & \multicolumn{5}{c}{\texttt{Multi-step / Iterative Policies}$^\dagger$} & \multicolumn{3}{c}{\texttt{One-step Methods}} \\
\cmidrule(lr){2-4} \cmidrule(lr){5-9} \cmidrule(lr){10-12}
\texttt{Task} & \texttt{BC} & \texttt{IQL} & \texttt{ReBRAC} & \texttt{FAWAC} & \texttt{FBRAC} & \texttt{IFQL} & \texttt{DeFlow} & \texttt{SORL} & \texttt{\color{myblue}FQL} & \texttt{\color{mygreen}DROL(16)} & \texttt{\color{myteal}DROL}$^*$ \\
\midrule
\texttt{antmaze-large-navigate-singletask-task1-v0 (*)} &    $0$ {\tiny $\pm 0$} &    $48$ {\tiny $\pm 9$} &    $91$ {\tiny $\pm 10$} &    $1$ {\tiny $\pm 1$} &    $70$ {\tiny $\pm 20$} &    $24$ {\tiny $\pm 17$} &    $88$ {\tiny $\pm 6$} &    $93$ {\tiny $\pm 2$} &    $80$ {\tiny $\pm 8$} &    $\underline{\mathbf{98}}$ {\tiny $\pm 2$} &    $\underline{\mathbf{98}}$ {\tiny $\pm 2$} \\
\texttt{antmaze-large-navigate-singletask-task2-v0} &    $6$ {\tiny $\pm 3$} &    $42$ {\tiny $\pm 6$} &    $\mathbf{88}$ {\tiny $\pm 4$} &    $0$ {\tiny $\pm 1$} &    $35$ {\tiny $\pm 12$} &    $8$ {\tiny $\pm 3$} &    $71$ {\tiny $\pm 10$} &    $79$ {\tiny $\pm 5$} &    $57$ {\tiny $\pm 10$} &    $\underline{\mathbf{86}}$ {\tiny $\pm 7$} &    $\underline{\mathbf{84}}$ {\tiny $\pm 6$} \\
\texttt{antmaze-large-navigate-singletask-task3-v0} &    $29$ {\tiny $\pm 5$} &    $72$ {\tiny $\pm 7$} &    $51$ {\tiny $\pm 18$} &    $12$ {\tiny $\pm 4$} &    $83$ {\tiny $\pm 15$} &    $52$ {\tiny $\pm 17$} &    $88$ {\tiny $\pm 5$} &    $88$ {\tiny $\pm 10$} &    $\underline{\mathbf{93}}$ {\tiny $\pm 3$} &    $88$ {\tiny $\pm 5$} &    $\underline{\mathbf{90}}$ {\tiny $\pm 5$} \\
\texttt{antmaze-large-navigate-singletask-task4-v0} &    $8$ {\tiny $\pm 3$} &    $51$ {\tiny $\pm 9$} &    $84$ {\tiny $\pm 7$} &    $10$ {\tiny $\pm 3$} &    $37$ {\tiny $\pm 18$} &    $18$ {\tiny $\pm 8$} &    $77$ {\tiny $\pm 5$} &    $\mathbf{91}$ {\tiny $\pm 2$} &    $80$ {\tiny $\pm 4$} &    $84$ {\tiny $\pm 8$} &    $79$ {\tiny $\pm 14$} \\
\texttt{antmaze-large-navigate-singletask-task5-v0} &    $10$ {\tiny $\pm 3$} &    $54$ {\tiny $\pm 22$} &    $90$ {\tiny $\pm 2$} &    $9$ {\tiny $\pm 5$} &    $76$ {\tiny $\pm 8$} &    $38$ {\tiny $\pm 18$} &    $84$ {\tiny $\pm 3$} &    $\mathbf{95}$ {\tiny $\pm 0$} &    $83$ {\tiny $\pm 4$} &    $\underline{\mathbf{95}}$ {\tiny $\pm 4$} &    $\underline{\mathbf{93}}$ {\tiny $\pm 3$} \\
\midrule
\texttt{antmaze-giant-navigate-singletask-task1-v0 (*)} &    $0$ {\tiny $\pm 0$} &    $0$ {\tiny $\pm 0$} &    $\mathbf{27}$ {\tiny $\pm 22$} &    $0$ {\tiny $\pm 0$} &    $0$ {\tiny $\pm 1$} &    $0$ {\tiny $\pm 0$} &    $4$ {\tiny $\pm 4$} &    $12$ {\tiny $\pm 6$} &    $4$ {\tiny $\pm 5$} &    $0$ {\tiny $\pm 0$} &    $\underline{\mathbf{28}}$ {\tiny $\pm 31$} \\
\texttt{antmaze-giant-navigate-singletask-task2-v0} &    $0$ {\tiny $\pm 0$} &    $1$ {\tiny $\pm 1$} &    $16$ {\tiny $\pm 17$} &    $0$ {\tiny $\pm 0$} &    $4$ {\tiny $\pm 7$} &    $0$ {\tiny $\pm 0$} &    $36$ {\tiny $\pm 13$} &    $0$ {\tiny $\pm 0$} &    $9$ {\tiny $\pm 7$} &    $0$ {\tiny $\pm 0$} &    $\underline{\mathbf{54}}$ {\tiny $\pm 39$} \\
\texttt{antmaze-giant-navigate-singletask-task3-v0} &    $0$ {\tiny $\pm 0$} &    $0$ {\tiny $\pm 0$} &    $34$ {\tiny $\pm 22$} &    $0$ {\tiny $\pm 0$} &    $0$ {\tiny $\pm 0$} &    $0$ {\tiny $\pm 0$} &    $2$ {\tiny $\pm 3$} &    $0$ {\tiny $\pm 0$} &    $0$ {\tiny $\pm 1$} &    $0$ {\tiny $\pm 0$} &    $\underline{\mathbf{77}}$ {\tiny $\pm 9$} \\
\texttt{antmaze-giant-navigate-singletask-task4-v0} &    $0$ {\tiny $\pm 0$} &    $0$ {\tiny $\pm 0$} &    $5$ {\tiny $\pm 12$} &    $0$ {\tiny $\pm 0$} &    $9$ {\tiny $\pm 4$} &    $0$ {\tiny $\pm 0$} &    $5$ {\tiny $\pm 8$} &    $\mathbf{25}$ {\tiny $\pm 18$} &    $14$ {\tiny $\pm 23$} &    $0$ {\tiny $\pm 0$} &    $8$ {\tiny $\pm 22$} \\
\texttt{antmaze-giant-navigate-singletask-task5-v0} &    $1$ {\tiny $\pm 1$} &    $19$ {\tiny $\pm 7$} &    $49$ {\tiny $\pm 22$} &    $0$ {\tiny $\pm 0$} &    $6$ {\tiny $\pm 10$} &    $13$ {\tiny $\pm 9$} &    $16$ {\tiny $\pm 19$} &    $6$ {\tiny $\pm 15$} &    $16$ {\tiny $\pm 28$} &    $13$ {\tiny $\pm 17$} &    $\underline{\mathbf{58}}$ {\tiny $\pm 26$} \\
\midrule
\texttt{humanoidmaze-medium-navigate-singletask-task1-v0 (*)} &    $1$ {\tiny $\pm 0$} &    $32$ {\tiny $\pm 7$} &    $16$ {\tiny $\pm 9$} &    $6$ {\tiny $\pm 2$} &    $25$ {\tiny $\pm 8$} &    $\mathbf{69}$ {\tiny $\pm 19$} &    $21$ {\tiny $\pm 11$} &    $\mathbf{67}$ {\tiny $\pm 4$} &    $19$ {\tiny $\pm 12$} &    $58$ {\tiny $\pm 15$} &    $55$ {\tiny $\pm 16$} \\
\texttt{humanoidmaze-medium-navigate-singletask-task2-v0} &    $1$ {\tiny $\pm 0$} &    $41$ {\tiny $\pm 9$} &    $18$ {\tiny $\pm 16$} &    $40$ {\tiny $\pm 2$} &    $76$ {\tiny $\pm 10$} &    $85$ {\tiny $\pm 11$} &    $69$ {\tiny $\pm 5$} &    $89$ {\tiny $\pm 3$} &    $\underline{\mathbf{94}}$ {\tiny $\pm 3$} &    $88$ {\tiny $\pm 24$} &    $\underline{\mathbf{95}}$ {\tiny $\pm 11$} \\
\texttt{humanoidmaze-medium-navigate-singletask-task3-v0} &    $6$ {\tiny $\pm 2$} &    $25$ {\tiny $\pm 5$} &    $36$ {\tiny $\pm 13$} &    $19$ {\tiny $\pm 2$} &    $27$ {\tiny $\pm 11$} &    $49$ {\tiny $\pm 49$} &    $63$ {\tiny $\pm 14$} &    $\mathbf{83}$ {\tiny $\pm 4$} &    $74$ {\tiny $\pm 18$} &    $78$ {\tiny $\pm 25$} &    $\underline{\mathbf{79}}$ {\tiny $\pm 20$} \\
\texttt{humanoidmaze-medium-navigate-singletask-task4-v0} &    $0$ {\tiny $\pm 0$} &    $0$ {\tiny $\pm 1$} &    $15$ {\tiny $\pm 16$} &    $1$ {\tiny $\pm 1$} &    $1$ {\tiny $\pm 2$} &    $1$ {\tiny $\pm 1$} &    $6$ {\tiny $\pm 7$} &    $1$ {\tiny $\pm 0$} &    $3$ {\tiny $\pm 4$} &    $15$ {\tiny $\pm 8$} &    $\underline{\mathbf{17}}$ {\tiny $\pm 10$} \\
\texttt{humanoidmaze-medium-navigate-singletask-task5-v0} &    $2$ {\tiny $\pm 1$} &    $66$ {\tiny $\pm 4$} &    $24$ {\tiny $\pm 20$} &    $31$ {\tiny $\pm 7$} &    $63$ {\tiny $\pm 9$} &    $\mathbf{98}$ {\tiny $\pm 2$} &    $81$ {\tiny $\pm 7$} &    $81$ {\tiny $\pm 20$} &    $\underline{\mathbf{97}}$ {\tiny $\pm 2$} &    $84$ {\tiny $\pm 24$} &    $\underline{\mathbf{96}}$ {\tiny $\pm 18$} \\
\midrule
\texttt{humanoidmaze-large-navigate-singletask-task1-v0 (*)} &    $0$ {\tiny $\pm 0$} &    $3$ {\tiny $\pm 1$} &    $2$ {\tiny $\pm 1$} &    $0$ {\tiny $\pm 0$} &    $0$ {\tiny $\pm 1$} &    $6$ {\tiny $\pm 2$} &    $3$ {\tiny $\pm 4$} &    $\mathbf{20}$ {\tiny $\pm 9$} &    $7$ {\tiny $\pm 6$} &    $\underline{\mathbf{21}}$ {\tiny $\pm 22$} &    $\underline{\mathbf{20}}$ {\tiny $\pm 15$} \\
\texttt{humanoidmaze-large-navigate-singletask-task2-v0} &    $\mathbf{0}$ {\tiny $\pm 0$} &    $\mathbf{0}$ {\tiny $\pm 0$} &    $\mathbf{0}$ {\tiny $\pm 0$} &    $\mathbf{0}$ {\tiny $\pm 0$} &    $\mathbf{0}$ {\tiny $\pm 0$} &    $\mathbf{0}$ {\tiny $\pm 0$} &    $\mathbf{0}$ {\tiny $\pm 4$} &    $\mathbf{0}$ {\tiny $\pm 0$} &    $\underline{\mathbf{0}}$ {\tiny $\pm 0$} &    $\underline{\mathbf{0}}$ {\tiny $\pm 0$} &    $\underline{\mathbf{0}}$ {\tiny $\pm 0$} \\
\texttt{humanoidmaze-large-navigate-singletask-task3-v0} &    $1$ {\tiny $\pm 1$} &    $7$ {\tiny $\pm 3$} &    $8$ {\tiny $\pm 4$} &    $1$ {\tiny $\pm 1$} &    $10$ {\tiny $\pm 2$} &    $\mathbf{48}$ {\tiny $\pm 10$} &    $10$ {\tiny $\pm 6$} &    $5$ {\tiny $\pm 2$} &    $11$ {\tiny $\pm 7$} &    $16$ {\tiny $\pm 7$} &    $34$ {\tiny $\pm 14$} \\
\texttt{humanoidmaze-large-navigate-singletask-task4-v0} &    $1$ {\tiny $\pm 0$} &    $1$ {\tiny $\pm 0$} &    $1$ {\tiny $\pm 1$} &    $0$ {\tiny $\pm 0$} &    $0$ {\tiny $\pm 0$} &    $1$ {\tiny $\pm 1$} &    $\mathbf{8}$ {\tiny $\pm 3$} &    $0$ {\tiny $\pm 0$} &    $2$ {\tiny $\pm 3$} &    $0$ {\tiny $\pm 0$} &    $1$ {\tiny $\pm 1$} \\
\texttt{humanoidmaze-large-navigate-singletask-task5-v0} &    $0$ {\tiny $\pm 1$} &    $1$ {\tiny $\pm 1$} &    $\mathbf{2}$ {\tiny $\pm 2$} &    $0$ {\tiny $\pm 0$} &    $1$ {\tiny $\pm 1$} &    $0$ {\tiny $\pm 0$} &    $1$ {\tiny $\pm 1$} &    $0$ {\tiny $\pm 0$} &    $1$ {\tiny $\pm 3$} &    $0$ {\tiny $\pm 0$} &    $0$ {\tiny $\pm 0$} \\
\midrule
\texttt{antsoccer-arena-navigate-singletask-task1-v0} &    $2$ {\tiny $\pm 1$} &    $14$ {\tiny $\pm 5$} &    $0$ {\tiny $\pm 0$} &    $22$ {\tiny $\pm 2$} &    $17$ {\tiny $\pm 3$} &    $61$ {\tiny $\pm 25$} &    $84$ {\tiny $\pm 7$} &    $\mathbf{93}$ {\tiny $\pm 4$} &    $77$ {\tiny $\pm 4$} &    $80$ {\tiny $\pm 6$} &    $76$ {\tiny $\pm 6$} \\
\texttt{antsoccer-arena-navigate-singletask-task2-v0} &    $2$ {\tiny $\pm 2$} &    $17$ {\tiny $\pm 7$} &    $0$ {\tiny $\pm 1$} &    $8$ {\tiny $\pm 1$} &    $8$ {\tiny $\pm 2$} &    $75$ {\tiny $\pm 3$} &    $89$ {\tiny $\pm 4$} &    $\mathbf{96}$ {\tiny $\pm 2$} &    $88$ {\tiny $\pm 3$} &    $\underline{\mathbf{93}}$ {\tiny $\pm 4$} &    $87$ {\tiny $\pm 7$} \\
\texttt{antsoccer-arena-navigate-singletask-task3-v0} &    $0$ {\tiny $\pm 0$} &    $6$ {\tiny $\pm 4$} &    $0$ {\tiny $\pm 0$} &    $11$ {\tiny $\pm 5$} &    $16$ {\tiny $\pm 3$} &    $14$ {\tiny $\pm 22$} &    $56$ {\tiny $\pm 7$} &    $55$ {\tiny $\pm 6$} &    $\underline{\mathbf{61}}$ {\tiny $\pm 6$} &    $\underline{\mathbf{58}}$ {\tiny $\pm 3$} &    $\underline{\mathbf{60}}$ {\tiny $\pm 7$} \\
\texttt{antsoccer-arena-navigate-singletask-task4-v0 (*)} &    $1$ {\tiny $\pm 0$} &    $3$ {\tiny $\pm 2$} &    $0$ {\tiny $\pm 0$} &    $12$ {\tiny $\pm 3$} &    $24$ {\tiny $\pm 4$} &    $16$ {\tiny $\pm 9$} &    $43$ {\tiny $\pm 5$} &    $\mathbf{54}$ {\tiny $\pm 5$} &    $39$ {\tiny $\pm 6$} &    $39$ {\tiny $\pm 9$} &    $41$ {\tiny $\pm 7$} \\
\texttt{antsoccer-arena-navigate-singletask-task5-v0} &    $0$ {\tiny $\pm 0$} &    $2$ {\tiny $\pm 2$} &    $0$ {\tiny $\pm 0$} &    $9$ {\tiny $\pm 2$} &    $15$ {\tiny $\pm 4$} &    $0$ {\tiny $\pm 1$} &    $39$ {\tiny $\pm 9$} &    $47$ {\tiny $\pm 9$} &    $36$ {\tiny $\pm 9$} &    $42$ {\tiny $\pm 12$} &    $\underline{\mathbf{60}}$ {\tiny $\pm 8$} \\
\midrule
\texttt{cube-single-play-singletask-task1-v0} &    $10$ {\tiny $\pm 5$} &    $88$ {\tiny $\pm 3$} &    $89$ {\tiny $\pm 5$} &    $81$ {\tiny $\pm 9$} &    $73$ {\tiny $\pm 33$} &    $79$ {\tiny $\pm 4$} &    $\mathbf{96}$ {\tiny $\pm 3$} &    $\mathbf{97}$ {\tiny $\pm 2$} &    $\underline{\mathbf{97}}$ {\tiny $\pm 2$} &    $\underline{\mathbf{97}}$ {\tiny $\pm 2$} &    $\underline{\mathbf{98}}$ {\tiny $\pm 2$} \\
\texttt{cube-single-play-singletask-task2-v0 (*)} &    $3$ {\tiny $\pm 1$} &    $85$ {\tiny $\pm 8$} &    $92$ {\tiny $\pm 4$} &    $81$ {\tiny $\pm 9$} &    $83$ {\tiny $\pm 13$} &    $73$ {\tiny $\pm 3$} &    $\mathbf{99}$ {\tiny $\pm 1$} &    $\mathbf{99}$ {\tiny $\pm 0$} &    $\underline{\mathbf{97}}$ {\tiny $\pm 2$} &    $\underline{\mathbf{99}}$ {\tiny $\pm 1$} &    $\underline{\mathbf{99}}$ {\tiny $\pm 1$} \\
\texttt{cube-single-play-singletask-task3-v0} &    $9$ {\tiny $\pm 3$} &    $91$ {\tiny $\pm 5$} &    $93$ {\tiny $\pm 3$} &    $87$ {\tiny $\pm 4$} &    $82$ {\tiny $\pm 12$} &    $88$ {\tiny $\pm 4$} &    $\mathbf{100}$ {\tiny $\pm 1$} &    $\mathbf{99}$ {\tiny $\pm 1$} &    $\underline{\mathbf{98}}$ {\tiny $\pm 2$} &    $\underline{\mathbf{99}}$ {\tiny $\pm 1$} &    $\underline{\mathbf{100}}$ {\tiny $\pm 1$} \\
\texttt{cube-single-play-singletask-task4-v0} &    $2$ {\tiny $\pm 1$} &    $73$ {\tiny $\pm 6$} &    $\mathbf{92}$ {\tiny $\pm 3$} &    $79$ {\tiny $\pm 6$} &    $79$ {\tiny $\pm 20$} &    $79$ {\tiny $\pm 6$} &    $\mathbf{95}$ {\tiny $\pm 4$} &    $\mathbf{95}$ {\tiny $\pm 2$} &    $\underline{\mathbf{94}}$ {\tiny $\pm 3$} &    $\underline{\mathbf{95}}$ {\tiny $\pm 3$} &    $\underline{\mathbf{96}}$ {\tiny $\pm 2$} \\
\texttt{cube-single-play-singletask-task5-v0} &    $3$ {\tiny $\pm 3$} &    $78$ {\tiny $\pm 9$} &    $87$ {\tiny $\pm 8$} &    $78$ {\tiny $\pm 10$} &    $76$ {\tiny $\pm 33$} &    $77$ {\tiny $\pm 7$} &    $\mathbf{92}$ {\tiny $\pm 5$} &    $\mathbf{93}$ {\tiny $\pm 3$} &    $\underline{\mathbf{93}}$ {\tiny $\pm 3$} &    $\underline{\mathbf{94}}$ {\tiny $\pm 5$} &    $\underline{\mathbf{96}}$ {\tiny $\pm 2$} \\
\midrule
\texttt{cube-double-play-singletask-task1-v0} &    $8$ {\tiny $\pm 3$} &    $27$ {\tiny $\pm 5$} &    $45$ {\tiny $\pm 6$} &    $21$ {\tiny $\pm 7$} &    $47$ {\tiny $\pm 11$} &    $35$ {\tiny $\pm 9$} &    $65$ {\tiny $\pm 6$} &    $\mathbf{77}$ {\tiny $\pm 11$} &    $61$ {\tiny $\pm 9$} &    $65$ {\tiny $\pm 7$} &    $65$ {\tiny $\pm 14$} \\
\texttt{cube-double-play-singletask-task2-v0 (*)} &    $0$ {\tiny $\pm 0$} &    $1$ {\tiny $\pm 1$} &    $7$ {\tiny $\pm 3$} &    $2$ {\tiny $\pm 1$} &    $22$ {\tiny $\pm 12$} &    $9$ {\tiny $\pm 5$} &    $\mathbf{49}$ {\tiny $\pm 13$} &    $33$ {\tiny $\pm 8$} &    $36$ {\tiny $\pm 6$} &    $22$ {\tiny $\pm 8$} &    $28$ {\tiny $\pm 7$} \\
\texttt{cube-double-play-singletask-task3-v0} &    $0$ {\tiny $\pm 0$} &    $0$ {\tiny $\pm 0$} &    $4$ {\tiny $\pm 1$} &    $1$ {\tiny $\pm 1$} &    $4$ {\tiny $\pm 2$} &    $8$ {\tiny $\pm 5$} &    $\mathbf{34}$ {\tiny $\pm 15$} &    $12$ {\tiny $\pm 6$} &    $22$ {\tiny $\pm 5$} &    $9$ {\tiny $\pm 5$} &    $22$ {\tiny $\pm 6$} \\
\texttt{cube-double-play-singletask-task4-v0} &    $0$ {\tiny $\pm 0$} &    $0$ {\tiny $\pm 0$} &    $1$ {\tiny $\pm 1$} &    $0$ {\tiny $\pm 0$} &    $0$ {\tiny $\pm 1$} &    $1$ {\tiny $\pm 1$} &    $7$ {\tiny $\pm 4$} &    $7$ {\tiny $\pm 4$} &    $5$ {\tiny $\pm 2$} &    $2$ {\tiny $\pm 2$} &    $\underline{\mathbf{9}}$ {\tiny $\pm 4$} \\
\texttt{cube-double-play-singletask-task5-v0} &    $0$ {\tiny $\pm 0$} &    $4$ {\tiny $\pm 3$} &    $4$ {\tiny $\pm 2$} &    $2$ {\tiny $\pm 1$} &    $2$ {\tiny $\pm 2$} &    $17$ {\tiny $\pm 6$} &    $\mathbf{46}$ {\tiny $\pm 11$} &    $1$ {\tiny $\pm 1$} &    $19$ {\tiny $\pm 10$} &    $2$ {\tiny $\pm 2$} &    $8$ {\tiny $\pm 2$} \\
\midrule
\texttt{scene-play-singletask-task1-v0} &    $19$ {\tiny $\pm 6$} &    $94$ {\tiny $\pm 3$} &    $\mathbf{95}$ {\tiny $\pm 2$} &    $87$ {\tiny $\pm 8$} &    $\mathbf{96}$ {\tiny $\pm 8$} &    $\mathbf{98}$ {\tiny $\pm 3$} &    $\mathbf{98}$ {\tiny $\pm 2$} &    $\mathbf{99}$ {\tiny $\pm 1$} &    $\underline{\mathbf{100}}$ {\tiny $\pm 0$} &    $\underline{\mathbf{100}}$ {\tiny $\pm 0$} &    $\underline{\mathbf{100}}$ {\tiny $\pm 0$} \\
\texttt{scene-play-singletask-task2-v0 (*)} &    $1$ {\tiny $\pm 1$} &    $12$ {\tiny $\pm 3$} &    $50$ {\tiny $\pm 13$} &    $18$ {\tiny $\pm 8$} &    $46$ {\tiny $\pm 10$} &    $0$ {\tiny $\pm 0$} &    $59$ {\tiny $\pm 17$} &    $\mathbf{89}$ {\tiny $\pm 9$} &    $76$ {\tiny $\pm 9$} &    $69$ {\tiny $\pm 12$} &    $\underline{\mathbf{89}}$ {\tiny $\pm 8$} \\
\texttt{scene-play-singletask-task3-v0} &    $1$ {\tiny $\pm 1$} &    $32$ {\tiny $\pm 7$} &    $55$ {\tiny $\pm 16$} &    $38$ {\tiny $\pm 9$} &    $78$ {\tiny $\pm 14$} &    $54$ {\tiny $\pm 19$} &    $85$ {\tiny $\pm 5$} &    $\mathbf{97}$ {\tiny $\pm 1$} &    $\underline{\mathbf{98}}$ {\tiny $\pm 1$} &    $\underline{\mathbf{98}}$ {\tiny $\pm 2$} &    $\underline{\mathbf{98}}$ {\tiny $\pm 2$} \\
\texttt{scene-play-singletask-task4-v0} &    $2$ {\tiny $\pm 2$} &    $0$ {\tiny $\pm 1$} &    $3$ {\tiny $\pm 3$} &    $6$ {\tiny $\pm 1$} &    $4$ {\tiny $\pm 4$} &    $0$ {\tiny $\pm 0$} &    $\mathbf{13}$ {\tiny $\pm 8$} &    $1$ {\tiny $\pm 1$} &    $5$ {\tiny $\pm 1$} &    $3$ {\tiny $\pm 4$} &    $3$ {\tiny $\pm 2$} \\
\texttt{scene-play-singletask-task5-v0} &    $\mathbf{0}$ {\tiny $\pm 0$} &    $\mathbf{0}$ {\tiny $\pm 0$} &    $\mathbf{0}$ {\tiny $\pm 0$} &    $\mathbf{0}$ {\tiny $\pm 0$} &    $\mathbf{0}$ {\tiny $\pm 0$} &    $\mathbf{0}$ {\tiny $\pm 0$} &    $\mathbf{0}$ {\tiny $\pm 0$} &    $\mathbf{0}$ {\tiny $\pm 0$} &    $\underline{\mathbf{0}}$ {\tiny $\pm 0$} &    $\underline{\mathbf{0}}$ {\tiny $\pm 0$} &    $\underline{\mathbf{0}}$ {\tiny $\pm 0$} \\
\midrule
\texttt{puzzle-3x3-play-singletask-task1-v0} &    $5$ {\tiny $\pm 2$} &    $33$ {\tiny $\pm 6$} &    $\mathbf{97}$ {\tiny $\pm 4$} &    $25$ {\tiny $\pm 9$} &    $63$ {\tiny $\pm 19$} &    $\mathbf{94}$ {\tiny $\pm 3$} &    $89$ {\tiny $\pm 7$} &    $-$ &    $90$ {\tiny $\pm 4$} &    $\underline{\mathbf{93}}$ {\tiny $\pm 9$} &    $\underline{\mathbf{93}}$ {\tiny $\pm 9$} \\
\texttt{puzzle-3x3-play-singletask-task2-v0} &    $1$ {\tiny $\pm 1$} &    $4$ {\tiny $\pm 3$} &    $1$ {\tiny $\pm 1$} &    $4$ {\tiny $\pm 2$} &    $2$ {\tiny $\pm 2$} &    $1$ {\tiny $\pm 2$} &    $\mathbf{38}$ {\tiny $\pm 11$} &    $-$ &    $16$ {\tiny $\pm 5$} &    $24$ {\tiny $\pm 31$} &    $24$ {\tiny $\pm 31$} \\
\texttt{puzzle-3x3-play-singletask-task3-v0} &    $1$ {\tiny $\pm 1$} &    $3$ {\tiny $\pm 2$} &    $3$ {\tiny $\pm 1$} &    $1$ {\tiny $\pm 0$} &    $1$ {\tiny $\pm 1$} &    $0$ {\tiny $\pm 0$} &    $\mathbf{35}$ {\tiny $\pm 11$} &    $-$ &    $10$ {\tiny $\pm 3$} &    $7$ {\tiny $\pm 4$} &    $7$ {\tiny $\pm 4$} \\
\texttt{puzzle-3x3-play-singletask-task4-v0 (*)} &    $1$ {\tiny $\pm 1$} &    $2$ {\tiny $\pm 1$} &    $2$ {\tiny $\pm 1$} &    $1$ {\tiny $\pm 1$} &    $2$ {\tiny $\pm 2$} &    $0$ {\tiny $\pm 0$} &    $23$ {\tiny $\pm 8$} &    $-$ &    $16$ {\tiny $\pm 5$} &    $\underline{\mathbf{36}}$ {\tiny $\pm 19$} &    $\underline{\mathbf{36}}$ {\tiny $\pm 19$} \\
\texttt{puzzle-3x3-play-singletask-task5-v0} &    $1$ {\tiny $\pm 0$} &    $3$ {\tiny $\pm 2$} &    $5$ {\tiny $\pm 3$} &    $1$ {\tiny $\pm 1$} &    $2$ {\tiny $\pm 2$} &    $0$ {\tiny $\pm 0$} &    $31$ {\tiny $\pm 8$} &    $-$ &    $16$ {\tiny $\pm 3$} &    $\underline{\mathbf{46}}$ {\tiny $\pm 23$} &    $\underline{\mathbf{46}}$ {\tiny $\pm 23$} \\
\midrule
\texttt{puzzle-4x4-play-singletask-task1-v0} &    $1$ {\tiny $\pm 1$} &    $12$ {\tiny $\pm 2$} &    $26$ {\tiny $\pm 4$} &    $1$ {\tiny $\pm 2$} &    $32$ {\tiny $\pm 9$} &    $\mathbf{49}$ {\tiny $\pm 9$} &    $23$ {\tiny $\pm 3$} &    $-$ &    $34$ {\tiny $\pm 8$} &    $28$ {\tiny $\pm 6$} &    $39$ {\tiny $\pm 6$} \\
\texttt{puzzle-4x4-play-singletask-task2-v0} &    $0$ {\tiny $\pm 0$} &    $7$ {\tiny $\pm 4$} &    $12$ {\tiny $\pm 4$} &    $0$ {\tiny $\pm 1$} &    $5$ {\tiny $\pm 3$} &    $4$ {\tiny $\pm 4$} &    $9$ {\tiny $\pm 2$} &    $-$ &    $16$ {\tiny $\pm 5$} &    $11$ {\tiny $\pm 3$} &    $\underline{\mathbf{20}}$ {\tiny $\pm 7$} \\
\texttt{puzzle-4x4-play-singletask-task3-v0} &    $0$ {\tiny $\pm 0$} &    $9$ {\tiny $\pm 3$} &    $15$ {\tiny $\pm 3$} &    $1$ {\tiny $\pm 1$} &    $20$ {\tiny $\pm 10$} &    $\mathbf{50}$ {\tiny $\pm 14$} &    $9$ {\tiny $\pm 4$} &    $-$ &    $18$ {\tiny $\pm 5$} &    $18$ {\tiny $\pm 7$} &    $32$ {\tiny $\pm 8$} \\
\texttt{puzzle-4x4-play-singletask-task4-v0 (*)} &    $0$ {\tiny $\pm 0$} &    $5$ {\tiny $\pm 2$} &    $10$ {\tiny $\pm 3$} &    $0$ {\tiny $\pm 0$} &    $5$ {\tiny $\pm 1$} &    $\mathbf{21}$ {\tiny $\pm 11$} &    $7$ {\tiny $\pm 4$} &    $-$ &    $11$ {\tiny $\pm 3$} &    $12$ {\tiny $\pm 4$} &    $16$ {\tiny $\pm 4$} \\
\texttt{puzzle-4x4-play-singletask-task5-v0} &    $0$ {\tiny $\pm 0$} &    $4$ {\tiny $\pm 1$} &    $7$ {\tiny $\pm 3$} &    $0$ {\tiny $\pm 1$} &    $4$ {\tiny $\pm 3$} &    $2$ {\tiny $\pm 2$} &    $7$ {\tiny $\pm 4$} &    $-$ &    $7$ {\tiny $\pm 3$} &    $12$ {\tiny $\pm 3$} &    $\underline{\mathbf{17}}$ {\tiny $\pm 4$} \\
\midrule
\texttt{antmaze-umaze-v2} &    $55$ &    $77$ &    $\mathbf{98}$ &    $90$ {\tiny $\pm 6$} &    $\mathbf{94}$ {\tiny $\pm 3$} &    $92$ {\tiny $\pm 6$} &    $\mathbf{96}$ {\tiny $\pm 3$} &    $-$ &    $\underline{\mathbf{96}}$ {\tiny $\pm 2$} &    $\underline{\mathbf{94}}$ {\tiny $\pm 2$} &    $\underline{\mathbf{94}}$ {\tiny $\pm 2$} \\
\texttt{antmaze-umaze-diverse-v2} &    $47$ &    $54$ &    $84$ &    $55$ {\tiny $\pm 7$} &    $82$ {\tiny $\pm 9$} &    $62$ {\tiny $\pm 12$} &    $\mathbf{87}$ {\tiny $\pm 8$} &    $-$ &    $\underline{\mathbf{89}}$ {\tiny $\pm 5$} &    $\underline{\mathbf{86}}$ {\tiny $\pm 8$} &    $\underline{\mathbf{86}}$ {\tiny $\pm 8$} \\
\texttt{antmaze-medium-play-v2} &    $0$ &    $66$ &    $\mathbf{90}$ &    $52$ {\tiny $\pm 12$} &    $77$ {\tiny $\pm 7$} &    $56$ {\tiny $\pm 15$} &    $80$ {\tiny $\pm 7$} &    $-$ &    $78$ {\tiny $\pm 7$} &    $84$ {\tiny $\pm 7$} &    $84$ {\tiny $\pm 7$} \\
\texttt{antmaze-medium-diverse-v2} &    $1$ &    $74$ &    $\mathbf{84}$ &    $44$ {\tiny $\pm 15$} &    $77$ {\tiny $\pm 6$} &    $60$ {\tiny $\pm 25$} &    $76$ {\tiny $\pm 19$} &    $-$ &    $71$ {\tiny $\pm 13$} &    $72$ {\tiny $\pm 11$} &    $72$ {\tiny $\pm 11$} \\
\texttt{antmaze-large-play-v2} &    $0$ &    $42$ &    $52$ &    $10$ {\tiny $\pm 6$} &    $32$ {\tiny $\pm 21$} &    $55$ {\tiny $\pm 9$} &    $78$ {\tiny $\pm 6$} &    $-$ &    $\underline{\mathbf{84}}$ {\tiny $\pm 7$} &    $64$ {\tiny $\pm 11$} &    $72$ {\tiny $\pm 4$} \\
\texttt{antmaze-large-diverse-v2} &    $0$ &    $30$ &    $64$ &    $16$ {\tiny $\pm 10$} &    $20$ {\tiny $\pm 17$} &    $64$ {\tiny $\pm 8$} &    $78$ {\tiny $\pm 6$} &    $-$ &    $\underline{\mathbf{83}}$ {\tiny $\pm 4$} &    $70$ {\tiny $\pm 7$} &    $76$ {\tiny $\pm 4$} \\
\midrule
\texttt{pen-human-v1} &    $71$ &    $78$ &    $\mathbf{103}$ &    $67$ {\tiny $\pm 5$} &    $77$ {\tiny $\pm 7$} &    $71$ {\tiny $\pm 12$} &    $61$ {\tiny $\pm 9$} &    $-$ &    $53$ {\tiny $\pm 6$} &    $52$ {\tiny $\pm 9$} &    $52$ {\tiny $\pm 4$} \\
\texttt{pen-cloned-v1} &    $52$ &    $83$ &    $\mathbf{103}$ &    $62$ {\tiny $\pm 10$} &    $67$ {\tiny $\pm 9$} &    $80$ {\tiny $\pm 11$} &    $75$ {\tiny $\pm 4$} &    $-$ &    $74$ {\tiny $\pm 11$} &    $78$ {\tiny $\pm 12$} &    $78$ {\tiny $\pm 12$} \\
\texttt{pen-expert-v1} &    $110$ &    $128$ &    $\mathbf{152}$ &    $118$ {\tiny $\pm 6$} &    $119$ {\tiny $\pm 7$} &    $139$ {\tiny $\pm 5$} &    $138$ {\tiny $\pm 5$} &    $-$ &    $142$ {\tiny $\pm 6$} &    $137$ {\tiny $\pm 6$} &    $137$ {\tiny $\pm 6$} \\
\midrule
\texttt{door-human-v1} &    $2$ &    $3$ &    $0$ &    $2$ {\tiny $\pm 1$} &    $4$ {\tiny $\pm 2$} &    $\mathbf{7}$ {\tiny $\pm 2$} &    $1$ {\tiny $\pm 0$} &    $-$ &    $0$ {\tiny $\pm 0$} &    $0$ {\tiny $\pm 0$} &    $0$ {\tiny $\pm 0$} \\
\texttt{door-cloned-v1} &    $0$ &    $\mathbf{3}$ &    $0$ &    $0$ {\tiny $\pm 1$} &    $0$ {\tiny $\pm 0$} &    $2$ {\tiny $\pm 2$} &    $2$ {\tiny $\pm 1$} &    $-$ &    $2$ {\tiny $\pm 1$} &    $1$ {\tiny $\pm 1$} &    $1$ {\tiny $\pm 1$} \\
\texttt{door-expert-v1} &    $\mathbf{105}$ &    $\mathbf{107}$ &    $\mathbf{106}$ &    $\mathbf{103}$ {\tiny $\pm 1$} &    $\mathbf{104}$ {\tiny $\pm 1$} &    $\mathbf{104}$ {\tiny $\pm 2$} &    $\mathbf{104}$ {\tiny $\pm 1$} &    $-$ &    $\underline{\mathbf{104}}$ {\tiny $\pm 1$} &    $\underline{\mathbf{105}}$ {\tiny $\pm 1$} &    $\underline{\mathbf{105}}$ {\tiny $\pm 0$} \\
\midrule
\texttt{hammer-human-v1} &    $\mathbf{3}$ &    $2$ &    $0$ &    $2$ {\tiny $\pm 1$} &    $2$ {\tiny $\pm 1$} &    $\mathbf{3}$ {\tiny $\pm 1$} &    $1$ {\tiny $\pm 0$} &    $-$ &    $1$ {\tiny $\pm 1$} &    $2$ {\tiny $\pm 2$} &    $2$ {\tiny $\pm 2$} \\
\texttt{hammer-cloned-v1} &    $1$ &    $2$ &    $5$ &    $1$ {\tiny $\pm 0$} &    $2$ {\tiny $\pm 1$} &    $2$ {\tiny $\pm 1$} &    $7$ {\tiny $\pm 4$} &    $-$ &    $\underline{\mathbf{11}}$ {\tiny $\pm 9$} &    $8$ {\tiny $\pm 3$} &    $8$ {\tiny $\pm 3$} \\
\texttt{hammer-expert-v1} &    $127$ &    $\mathbf{129}$ &    $\mathbf{134}$ &    $118$ {\tiny $\pm 3$} &    $119$ {\tiny $\pm 9$} &    $117$ {\tiny $\pm 9$} &    $127$ {\tiny $\pm 1$} &    $-$ &    $125$ {\tiny $\pm 3$} &    $\underline{\mathbf{128}}$ {\tiny $\pm 1$} &    $\underline{\mathbf{128}}$ {\tiny $\pm 1$} \\
\midrule
\texttt{relocate-human-v1} &    $\mathbf{0}$ &    $\mathbf{0}$ &    $\mathbf{0}$ &    $\mathbf{0}$ {\tiny $\pm 0$} &    $\mathbf{0}$ {\tiny $\pm 0$} &    $\mathbf{0}$ {\tiny $\pm 0$} &    $\mathbf{0}$ {\tiny $\pm 0$} &    $-$ &    $\underline{\mathbf{0}}$ {\tiny $\pm 0$} &    $\underline{\mathbf{0}}$ {\tiny $\pm 0$} &    $\underline{\mathbf{0}}$ {\tiny $\pm 0$} \\
\texttt{relocate-cloned-v1} &    $0$ &    $0$ &    $\mathbf{2}$ &    $0$ {\tiny $\pm 0$} &    $1$ {\tiny $\pm 1$} &    $0$ {\tiny $\pm 0$} &    $1$ {\tiny $\pm 8$} &    $-$ &    $0$ {\tiny $\pm 0$} &    $0$ {\tiny $\pm 0$} &    $0$ {\tiny $\pm 0$} \\
\texttt{relocate-expert-v1} &    $\mathbf{108}$ &    $\mathbf{106}$ &    $\mathbf{108}$ &    $\mathbf{105}$ {\tiny $\pm 3$} &    $\mathbf{105}$ {\tiny $\pm 2$} &    $\mathbf{104}$ {\tiny $\pm 3$} &    $\mathbf{109}$ {\tiny $\pm 2$} &    $-$ &    $\underline{\mathbf{107}}$ {\tiny $\pm 1$} &    $\underline{\mathbf{107}}$ {\tiny $\pm 2$} &    $\underline{\mathbf{108}}$ {\tiny $\pm 1$} \\
\bottomrule
\end{tabular}
\end{threeparttable}
\hspace{0pt}}
\vspace{-10pt}
\end{table*}

\clearpage